\documentclass[journal,article,submit,moreauthors,pdftex,10pt,a4paper]{mdpi} 
\pdfoutput=1
\firstpage{1} 
\pubvolume{xx}
\issuenum{1}
\articlenumber{5}
\pubyear{2018}
\copyrightyear{2018}
\preto{\abstractkeywords}{\nolinenumbers}
\Title{Geometric Estimation of Multivariate Dependency}

\Author{Salimeh~Yasaei Sekeh and Alfred~O.~Hero}

\AuthorNames{Salimeh Yasaei Sekeh and Alfred O. Hero}

\address{Department of Electrical Engineering and Computer Science, University of Michigan, Ann Arbor, USA; salimehy, hero@umich.edu}

\abstract{This paper proposes a geometric estimator of dependency between a pair of multivariate samples. The proposed estimator of dependency is based on a randomly permuted geometric graph (the minimal spanning tree) over the two multivariate samples. This estimator converges to a quantity that we call the geometric mutual information (GMI), which is equivalent to the Henze-Penrose divergence \cite{HP} between the joint distribution of the multivariate samples and the product of the marginals. The GMI has many of the same properties as standard MI but can be estimated from empirical data without density estimation; making it scalable to large datasets. The proposed empirical estimator of GMI is simple to implement, involving the construction of a MST spanning over both the original data and a randomly permuted version of this data. We establish asymptotic convergence of the estimator and convergence rates of the bias and variance for smooth multivariate density functions belonging to a H\"{o}lder class. We demonstrate the advantages of our proposed geometric dependency estimator in a series of experiments.}

\def\bom{{\mbox{\boldmath$\omega$}}}
\def\obom{\overline\bom}
\def\0bom{{\bom}^0}
\def\0obom{{\obom}^0}

\def\0nbom{{\bom}_{n,0}}
\def\n*bom{{\bom}^*_{(n)}}

\def\lam{\lambda}

\def\bbE{\mathbb E}

\def\bt{\mathbf t}

\def\bx{\mathbf x}

\def\bY{\mathbf Y}
\def\by{\mathbf y}

\def\bz{\mathbf z}

\def\bbB{\mathbb B}

\def\bbR{\mathbb R}
\def\bbS{\mathbb S}

\def\bx{\mathbf x}

\def\by{\mathbf y}

\def\bY{\mathbf Y}

\def\bx{\mathbf x}

\def\bz{\mathbf z}

\def\diy{\displaystyle}

\def\u0{{\underline 0}}

\def\rd{{\rm d}}

\def\rd{{\rm d}}

\def\bx{\mathbf x} \def\by{\mathbf y} \def\bz{\mathbf z}

\def\beq{\begin{eqnarray}} \def\eeq{\end{eqnarray}}
\def\beqq{\begin{eqnarray*}} \def\eeqq{\end{eqnarray*}}
\def\rd{{\rm d}} 
\def\BX{\mathbf{X}}\def\BY{\mathbf{Y}}
\def\BZ{\mathbf{Z}}

\def\1{{\mathbf 1}}   
\def\diy{\displaystyle} \def\bbE{{\mathbb E}} 
 \def\lam{\lambda} \def\bbB{{\mathbb B}}
\def\bbR{{\mathbb R}}\def\bbS{{\mathbb S}}

  \def\B1{\mathbf 1}

\def\BW{\mathbf{W}} 

\def\beal{\begin{array}{l}}
\def\beac{\begin{array}{c}}
\def\beacl{\begin{array}{cl}}
\def\ena{\end{array}}

\def\1{{\mathbf 1}}   
\def\diy{\displaystyle} \def\bbE{{\mathbb E}} 
 \def\lam{\lambda}
\def\bbB{{\mathbb B}} 
\def\bbR{{\mathbb R}}\def\bbS{{\mathbb S}}

  \def\B1{\mathbf 1}

 \def\BW{\mathbf{W}}

\def\beal{\begin{array}{l}}
\def\beac{\begin{array}{c}}
\def\beacl{\begin{array}{cl}}
\def\ena{\end{array}}

\def\beac{\begin{array}{c}} \def\beal{\begin{array}{l}} \def\beacl{\begin{array}{cl}} \def\ena{\end{array}}
\def\bbV{\mathbb{V}}
\def\bbS{\mathbb{S}}
\begin{document}

\section{Introduction}

Estimation of multivariate dependency has many applications in fields such as information theory, clustering, structure learning, data processing, feature selection, time series prediction and reinforcement learning, see \cite{LBP} and \cite{PHD}, \cite{MNYSH}, \cite{Bi}, \cite{Tor,VE,PLD,JZ}, \cite{SHL}, and \cite{MR}, respectively. It is difficult to accurately estimate the mutual information in high dimensional settings where the data is multivariate with a Lebesgue continuous density---the setting considered in this paper. An important and ubiquitous measure of dependency is the Shannon Mutual Information (MI), which has seen extensive use across many application domains. However, the estimation of mutual information can often be challenging.
%An especially attractive measure of dependency is the generalized mutual information (MI), which can be defined in terms of any $f$-divergence measure.
In this paper we focus on a measure of MI that we call the Geometric MI (GMI). This MI measure is defined as the asymptotic large sample limit of a randomized minimal spanning tree (MST) statistic spanning the multivariate sample realizations. The GMI is related to a divergence measure called the Henze-Penrose divergence \cite{neemuchwala2005entropic}, \cite{neemuchwala2006image}, and related to the multivariate runs test \cite{HP}. In \cite{BH}, \cite{BWHS}, it was shown that this divergence measure can be used to specify a tighter bound for the Bayes error rate for testing if a random sample comes from one of two distributions as compared to previous divergence-type bounds such as the Bhattacharrya bound. Furthermore, the authors of \cite{Yasaeietal2018} proposed a nonparametric bound on multi-class classification Bayes error rate using a global MST graph. 

Let $\BX$ and $\BY$ be random variables with unknown joint density $f_{{X}{Y}}$ and marginal densities $f_{{X}}$ and $f_{{Y}}$, respectively, and consider two hypotheses: $H_0$, $\BX$ and $\BY$ are independent and $H_1$, $\BX$ and $\BY$ are dependent,
\begin{equation*}
H_0: f_{{X}{Y}}=f_{{X}} f_{{Y}},\;\;\; \;\; \hbox{versus}\;\;\;\;\; H_1:f_{{X}{Y}} \neq f_{{X}} f_{{Y}}.
\end{equation*}
The GMI is defined as the Henze-Penrose divergence between $f_{XY}$ and $f_{X}f_{Y}$ which can be used as a dependency measure. In this paper we prove that for large sample size the randomized minimal spanning tree (MST) statistic spanning the original multivariate sample realizations and randomly shuffled data set in which for given the first element in each pair the second element is replaced by a randomly selected from all second elements, convergences to the GMI measure. A direct implication of \cite{BH}, \cite{BWHS} is that the GMI provides a tighter bound on the Bayes misclassification rate for the independence test of $H_0$ vs $H_1$. 
%In addition to the advantage of tighter bounds on the Bayes error rate, we have attempted to state another representation of GMI in terms of copula which provides an interesting application for nonparanormal family.
In this paper we propose an estimator based on a random permutation modification of the Friedman-Rafsky multivariate test statistic and show that under certain conditions the GMI estimator achieves the parametric mean square error (MSE) rate when the joint and marginal densities are bounded and smooth. Importantly our proposed GMI estimator does not require explicit estimation of the joint and marginal densities. 

In addition to achieving the optimal theoretical MSE rate, computational complexity is an important challenge addressed by researchers in machine learning and data science. Most plug-in based estimators like kernel density estimator (KDE) or K-nearest neighbour (KNN) estimator require runtime complexity of $O(n^2)$ which is not suitable for large scale applications. Noshad et al. proposed a graph theoretic direct estimation method based on nearest neighbor ratios (NNR) \cite{NMYH}. The NNR estimator is based on $k$-NN graph and computationally more tractable than other competing estimators with complexity $O(k n\log n)$. The construction of the minimal spanning tree lies at the heart of the GMI estimator proposed in this paper. Since the GMI estimator is based on the Euclidean MST the dual-tree algorithm by March et al. \cite{March2010} can be applied. This algorithm is based on the construction of Bor\r{u}vka \cite{Boruvka1926} and implements the Euclidean MST in approximately $O(n log n)$ time which is faster than the previous estimators. In this paper we experimentally show that for large number of sample sizes the proposed GMI estimator is less computationally complex than the KDE plug-in method.

\subsection{Related work}

Estimation of mutual information is related to estimation of the information histogram. The most common estimators of MI are based on plug-in density estimation, e.g., using kernel density or kNN density estimators \cite{KSG}, \cite{MSH2017}. Motivated by ensemble methods applied to divergence estimation \cite{MH}-\cite{MSGHcon}, in \cite{MSH2017} an ensemble method for combining multiple KDE bandwidths was proposed for estimating MI. Under certain smoothness conditions this  ensemble MI estimator was shown to achieve parametric convergence rates.  

Another class of estimators of multivariate dependency bypasses the difficult density estimation task. This class includes the statistically consistent estimator of R\'enyi-$\alpha$ and KL mutual information which is based on the asymptotic of the length of the KNN graph, \cite{LPS} and \cite{Gaoetal2015}. The estimator of \cite{PPS} builds on the KNN methods for R\'{e}nyi entropy estimation. As MI increases, the dependencies between random variables increase which results in less smooth densities. In \cite{GSG}, it has been shown that for large MI the KNN or KDE approaches are ill-suited candidates for estimating MI since the assumption of local uniformity can be violated when they are strong dependencies. To overcome this issue
an assumption on the smoothness of the density is required,
%the densities smoothness requirement is required, 
see \cite{Krishnamurthyetal2014}, \cite{Kandasamy2015}, and \cite{MH}-\cite{MSGHcon}. For all of these methods, the optimal parametric rate of MSE convergence is achieved when the densities are either $d$, $(d+1)/2$ or $d/2$ times differentiable \cite{SinghPoczos2016}. 
%Later the authors of \cite{SinghPoczos2016} improved the previous results and derived bounds on the bias of the KL estimate when the smoothness parameter is larger than $d/2$. 
In this paper, we assume that  joint and marginal densities are smooth in the sense that they belong to a strong H\"{o}lder continuous classes of densities $\Sigma^s_d(\eta,K)$, such that the smoothness parameter $\eta$ ranges in $(0,1]$. 

\vspace{0.2cm}
A MI measure based on the Pearson divergence was considered in \cite{Sugiyama2012} that is computational efficient and numerically stable. The authors of \cite{CostaHero2004} and \cite{Paletal2010} used minimal spanning tree generalized nearest-neighbor graph approaches, respectively, to estimate R\'{e}nyi mutual information. In \cite{MSH2017}, a nonparametric mutual information estimator was proposed using a weighted ensemble method with $O(1/n)$ parametric convergence rate. This estimator was based on plug-in density estimation, which is challenging in high dimension.

Our proposed estimator differs from previous methods in the following ways. First, it estimates a different measure of mutual information, the GMI. Second, instead of using the KNN graph the estimator of GMI uses a randomized minimal spanning tree that spans the multivariate realizations. The proposed GMI estimator is motivated by the multivariate runs test of Friedman and Rafsky (FR) \cite{FR} which is a multivariate generalization of the univariate Smirnov maximum deviation test \cite{Sm} and the Wald-Wolfowitz \cite{WW0} runs test in one dimension. We also emphasize that our proposed GMI estimator does not require boundary correction, in contrast to other graph-based estimators, such as, the NNR estimator \cite{NMYH}, scalable MI estimator \cite{NoshadHero2018}, or cross match statistic \cite{YOH}.

\subsection{Contribution}
The contribution of this paper has three components
\begin{itemize}
    \item [(1)] We propose a novel non-parametric multivariate dependency measure, referred to as Geometric Mutual Information (GMI), that is based on graph-based divergence estimation. The geometric mutual information is constructed using a minimal spanning tree and is a function of Friedman-Rafsky multivariate test statistic.
    \item [(2)] We establish properties of the proposed dependency measure analogous to those of standard mutual information, such as, convexity, concavity, chain rule, and the data processing inequality. 
    \item [(3)] We derive a bound on the MSE rate for the proposed geometric estimator. An advantage of the estimator is that it achieves the optimal MSE rate without explicitly performing boundary correction, which is required for most plug-in estimators.
\end{itemize}

\subsection{Organization}
The paper is organized as follows. In Section \ref{sec:HP-MI}, we define the geometric mutual information and establish some of its mathematical properties. In Subsection \ref{subsec:2.1} and \ref{Con:rates}, we introduce a statistically consistent GMI estimator and derive a bound on its mean square error convergence rate. In Section \ref{experiments} we verify the theory through experiments.

Throughout the paper, we denote statistical expectation by $\bbE$, the variance by abbreviation ${\rm Var}$. Bold face type indicates random vectors. All densities are assumed to be Lebesgue continuous with no atoms.

%\section{The geometric mutual information (GMI)}\label{sec:HP-MI}
\section{The geometric mutual information (GMI)}\label{sec:HP-MI}
In this section, we first review the definition of Henze-Penrose (HP) divergence measure defined by Berisha and Hero in \cite{BH} and \cite{HP}. The Henze-Penrose divergence between densities $f$ and $g$ with domain $\bbR^d$ for parameter $p\in(0,1)$ is defined as follows (see \cite{HP}, \cite{BerishaHero2015}, and \cite{Berishaetal2016}): 
\begin{equation} \label{EQ:DP} D_p(f,g)=\diy\frac{1}{4pq}\left[\int \frac{\big(p f(\bx)-q g(\bx)\big)^2}{p f(\bx)+q g(\bx)}\;\rd \bx -(p-q)^2\right],\end{equation}
where $q=1-p$. 
This functional is an $f$-divergence \cite{Cs}, also known as an Ali-Silvey distance \cite{AS}. That is, it satisfies the properties of non-negativity, monotonicity, and joint convexity \cite{BWHS}. This measure takes values in $[0,1]$ and $D_p=0$ if and only if $f=g$ almost surely. 

The geometric mutual information measure is defined as follows. Let $f_{X}$, $f_{Y}$, and $f_{X Y}$ be the marginal and joint distributions, respectively, of random vectors $\BX\in \bbR^{d_x}$, $\BY\in \bbR^{d_y}$ where $d_x$ and $d_y$ are positive integers. Then by using (\ref{EQ:DP}), a Henze-Penrose generalization of the mutual information between $\BX$ and $\BY$, is defined by
\begin{equation} \label{EQ:MIP}\begin{array}{l}I_p(\BX;\BY)=D_p(f_{XY},f_X f_Y)\\
\\
\qquad \qquad =\diy\frac{1}{4pq}\left[\iint \frac{\big(p  f_{XY}(\bx,\by)-q f_{X}(\by)f_Y(\by)\big)^2}{p f_{XY}(\bx,\by)+q f_{X}(\bx)f_Y(\by)}\;\rd \bx\; \rd\by -(p-q)^2\right].\end{array}\end{equation}
We will show below that $I_p(\BX;\BY)$ has a geometric interpretation in terms of the large sample limit of a minimal spanning tree spanning $n$ sample realizations of $\BX \cup \BY$. Thus we call $I_p(\BX;\BY)$ the geometric mutual information (GMI) between $\BX$ and $\BY$. The GMI satisfies similar properties to other definitions of mutual information, such as Shannon and R\'{e}nyi mutual information. Recalling (3) in \cite{BH}, an alternative form of $I_p$ is given by
\begin{align} I_p(\BX;\BY)=1-A_p(\BX;\BY)=\diy \frac{u_p(\BX;\BY)}{4 p q}-\diy \frac{(p-q)^2}{4 p q},\end{align}
where
\begin{equation}\begin{array}{cl}
A_p(\BX;\BY)=\diy \iint\frac{ f_{XY}(\bx,\by) f_X(\bx) f_{Y}(\by)}{pf_{XY}(\bx,\by) +q f_X(\bx)f_{Y}(\by)}\;\rd \bx\;\rd\by
=\diy\bbE_{XY}\bigg[\Big(p\;\frac{f_{XY}(\BX,\BY) }{f_X(\BX) f_{Y}(\BY)}+q\Big)^{-1}\bigg],\;\; \hbox{and}\\
\\
\diy u_p(\BX;\BY)=1-4 p q\; A_p(\BX;\BY).\end{array}\end{equation}
The affinity $A_p(\BX;\BY)$ is called the geometric affinity between $\BX$ and $\BY$, and it was originally defined in \cite{HP}.  The next subsection of the paper is dedicated to the basic inequalities and properties of the proposed GMI measure (\ref{EQ:MIP}).

\subsection{Properties of the geometric mutual information}\label{sec:prop}

In this subsection we establish basic inequalities and properties of the GMI, $I_p$, given in (\ref{EQ:MIP}). 
\subsubsection{Basic inequalities}
 The following theorem shows that $I_p(\BX;\bY)$ is a concave function in $f_{X}$ and a convex function in $f_{Y|X}$. The proof is given in Appendix A, Subsection \ref{conc.conx}. 

\def\lam{\lambda}
\begin{Theorem}\label{thm:convex/concave}
Denote by $\tilde{I}_p(f_{XY})$ the GMI $I_p(\BX;\BY)$ when $\BX\in \bbR^{d_x}$ and $\BY\in \bbR^{d_y}$ have joint density $f_{XY}$.  Then the GMI satisfies
\begin{itemize}
\item[] {\rm (i) Concavity in $f_X$}: Let $f_{Y|X}$ be conditional density of $\BY$ given $\BX$ and let $g_X$ and $h_X$ be densities on $\bbR^{d_x}$. For $p\in (0,1)$
\begin{equation}\label{eq1:thm.COn.PMI}\tilde{I}_p\big(\lambda_1 f_{Y|X}g_X+\lambda_2 f_{Y|X} h_X\big) \geq \lambda_1 \tilde{I}_p( f_{Y|X}g_X)+\lambda_2 \tilde{I}_p(f_{Y|X}h_X).\end{equation}
The inequality is strict unless either $\lambda_1$ or $\lambda_2$ are zero or $h_X=g_X$.

\item[] {\rm (ii) Convexity in $f_{Y|X}$}: Let $g_{Y|X}$ and $h_{Y|X}$ be conditional densities of $\BY$ given $\BX$ and let $f_X$ be marginal density. Then for $p\in (0,1)$
\begin{equation}\label{eq2:thm.2.1}
\tilde{I}_p\big(\lambda_1 g_{Y|X}f_X+\lambda_2 h_{Y|X} f_X\big) \leq \lambda_1 \tilde{I}_p( g_{Y|X}f_X)+\lambda_2 \tilde{I}_p(h_{Y|X}f_X) .
\end{equation}
The equality occurs when either $\lambda_1$ or $\lambda_2$ are zero or $h_{Y|X}=g_{Y|X}$.
\end{itemize}
\end{Theorem}
The GMI, $I_p(\BX;\BY)$, satisfies a property analogous to the standard chain rule and the data processing inequality \cite{TC}. For random variables $\BX\in \bbR^{d_x},\;\BY\in \bbR^{d_y}$, and $\BZ\in\bbR^{d_z}$ we define the conditional GMI measure by
\beqq\begin{array}{l} I_p(\BX;\BY|\BZ)=\diy\bbE_{\BZ}\Big[I_p(\BX;\BY|\BZ=\bz)\Big],\;\;\;\hbox{where}\\
\\
I_p(\BX;\BY|\BZ=\bz)=
\quad 1-\diy\iint \frac{f_{XY|Z}(\bx,\by|\bz) f_{X|Z}(\bx|\bz) f_{Y|Z}(\by|\bz)}{p\;f_{XY|Z}(\bx,\by|\bz)+q\;f_{X|Z}(\bx|\bz)f_{Y|Z}(\by|\bz)}\;\rd \bx\;\rd \by. \end{array}\eeqq
The next theorem establishes a relation between the joint and conditional geometric mutual information.
\def\BZ{\mathbf{Z}} \def\BY{\mathbf{Y}}\def\bz{\mathbf{z}}
\begin{Theorem}\label{thm:Chain.0}
For given $d$-dimensional random vector $\BX$ with components $X_1,X_2,\dots,X_d$ and random variable $Y$,
\begin{itemize}
\item [(i)] For $d=2$, set
\begin{equation}
    \delta_{X_2,Y|X_1}=\diy\bbE_f\left[\bigg(p\;\frac{f_{X_2Y|X_1}(x_2,y|x_1)}{f_{X_2|X_1}(x_2|x_1)\;f_{Y|X_1}(y|x_1)}+q\bigg)^{-1}\right].
\end{equation}
Here $f$ is the joint PDF of random vector $(X_1,X_2,Y)$ and note that $0\leq \delta_{X_2,Y|X_1}\leq 1$. Then we have 
\begin{equation}\label{eq1:thm22}
    I_p(X_1,X_2;Y)\geq I_p(X_1;Y)-\delta_{X_2,Y|X_1}.
\end{equation}
Furthermore $\delta_{X_i,Y|\BX^{i-1}}\in[0,1]$.
\bigskip
\item[(ii)] Let $\BX^i:=X_1,X_2,\dots,X_i$. Define a general form of $\delta$ in (i) by
\begin{equation}
    \delta_{X_i,Y|\BX^{i-1}}=\diy\bbE_f\bigg(p\;\frac{f_{X_iY|X^{i-1}}(x_i,y|x^{i-1})}{f_{X_i|X^{i-1}}(x_2|x^{i-1})\;f_{Y|X^{i-1}}(y|x^{i-1})}+q\bigg)^{-1},
\end{equation}
where $f$ is the join PDF of random vector $(X_1,\dots,X_d,Y)$. A general form of (\ref{eq1:thm22}) is given by 
\begin{equation}\label{eq2:thm2.2}
 I_p(\BX;Y)\geq I_p(X_1;Y)-\diy\sum_{i=1}^{d-1}\delta_{X_i,Y|\BX^{i-1}}.
\end{equation}
\end{itemize}
\end{Theorem}
Note that in the special case $\diy\sum_{i=1}^{d-1}\delta_{X_i,Y|\BX^{i-1}}\geq 1$, the inequality (\ref{eq2:thm2.2}) is trivial. 
The proof of Theorem \ref{thm:Chain.0} is given in Appendix B, subsection \ref{Proof:thm:Chain.0}. 
%------------------------------------------------------------
%\begin{theorem}\label{thm:Chain.I}
%For given $d$-dimensional random vector $\BX$ with components $X_1,X_2,\dots, X_d$ and random variable $Y$, we have
%\beq I_p(\BX;Y)\geq \sum\limits_{i=1}^d I_p(X_i;Y|X_{i-1},\dots,X_1)-(d-1). \eeq
%Furthermore, $X_i$ and $Y$ are conditionally independent given $X_{i-1},\dots,X_1$, for $i=1,\dots,d$ if and only if
%\beq  I_p(\BX;Y)\geq I_p(X_1;Y)-(d-1). \eeq
%\end{theorem}
%The proof of Theorem \ref{thm:Chain.I} is given in Appendix \ref{chainI}.
%---------------------------------------------------------
Next we apply Theorem \ref{thm:Chain.0} and state a result for Markov chain random vectors. Likewise to previous proofs, the proof of Proposition \ref{prop.1} is provided in Appendices section, Appendix C. 
\begin{Proposition}\label{prop.1}
Suppose random vectors $\mathbf{X},\BY,\BZ$ form a Markov chain denoted, $\BX\rightarrow \BY\rightarrow \BZ$, in the sense that $f_{XYZ}=f_{X|Y}f_{Y|Z}f_Z$. Then for $p\in(0,1)$
\beq\label{RHS:data.Proc.} I_p(\BY;\BX)\geq I_p(\BZ;\BX)-\Big(p\; \bbE_{XY}\big[\delta_{X,Y}\big]+q\Big)^{-1},\eeq
where
\vspace{-10pt}
\beqq \delta_{X,Y}=\diy\int\diy\frac{f_{X|Y}(\bx|\by)\;f_{Z|Y}(\bz|\by)}{f_{X|Z}(\bx|\bz)}\;\rd \bz. \eeqq
\end{Proposition}
Further, if both $\BX\rightarrow \BY\rightarrow \BZ$ and $\BX\rightarrow \BZ \rightarrow \BY$ together hold true, we have $I_p(\BY;\BX)=I_p(\BZ;\BX)$. \\
%Also, when both $\BX\rightarrow \BY\rightarrow \BZ$ and $\BY\rightarrow \BZ \rightarrow \BX$ hold true, we have $I_p(\BY;\BX)=I_p(\BZ;\BX)$. \\
The inequality in (\ref{RHS:data.Proc.}) becomes a tighter inequality interpretable as a standard data-processing inequality $I_p(\BY;\BX)\geq I_p(\BZ;\BX)$, when %$\bbE_{XY}\big[\delta_{X,Y}\big]$ does not exist, or specifically 
$$ \bbE_Z\left[\frac{f(\BZ|\BY)}{f(\BZ|\BX)}\right]=\infty, $$
since 
$$\bbE_{XY}\big[\delta_{X,Y}\big]=\bbE_{XY}\left(\diy\frac{f(\BX|\BY)}{f(\BX)} \bbE_Z\left[\frac{f(\BZ|\BY)}{f(\BZ|\BX)}\right]\right).$$
\subsection{The Friedman-Rafsky Estimator}\label{subsec:2.1}

Let a random sample $\{\bx_i, \by_i\}_{i=1}^{n}$ from $f_{XY}(\bx,\by)$ be available.  Here we show that the GMI $I_p(\BX;\BY)$ can be directly estimated without estimating the densities. The estimator is inspired by the MST construction of \cite{FR} that provides a consistent estimate of the Henze-Penrose divergence \cite{BH}, \cite{BWHS}. We denote by $\bz_i$ the $i$-th joint sample $\bx_i, \by_i$ and by $\mathcal{Z}_n$ the sample set $\{\bz_i\}_{i=1}^n$. Divide the sample set $\mathcal{Z}_n$ into two subsets $\mathcal{Z}'_{n'}$ and $\mathcal{Z}''_{n''}$ with the proportion $\alpha=n'/n$ and $\beta=n''/n$, where $\alpha+\beta=1$. 

Denote by $\widetilde{\mathcal{Z}}_{n''}$ the set
$$\big\{(\bx_{i_k},\by_{j_k}),\;k=1,\dots,n'',\; \hbox{selected at random from}\; {\mathcal{Z}}''_{n''}\big\}:$$ 
\vspace{-0.6cm}
\tikzset{every picture/.style={remember picture}}
\beqq \left(\begin{array}{c}
\tikz[baseline]{\node (a11) {$x_1$};}\\
\tikz[baseline]{\node (a21) {$x_2$};}\\
\tikz[baseline]{\node (a31) {$\vdots$};}\\
\tikz[baseline]{\node (a41) {$x_{n''}$};}
\end{array}\right)\qquad  \left(\begin{array}{c}
\tikz[baseline]{\node (a12) {$y_1$};}\\
\tikz[baseline]{\node (a22) {$y_2$};}\\
\tikz[baseline]{\node (a32) {$\vdots$};}\\
\tikz[baseline]{\node (a42) {$y_{n''}$};}
\end{array}\right)
\eeqq 
\begin{tikzpicture}[overlay]
    \path[thick,-] (a11) edge (a32);
    \path[thick,-] (a21) edge (a12);
    \path[thick,-] (a31) edge (a42);
    \path[thick,-] (a41) edge (a22);
\end{tikzpicture}
%This means that for each $\bz_{ik}=(\bx_{ik},\by_{ik})\in\mathcal{Z}''_{n''}$ given the first element $\bx_{ik}$ the second element $\by_{ik}$ is replaced by a second element of the selected randomly pairs from the set $\mathcal{Z}''_{n''}$ i.e. $\by_{jk}$ to generate the pair $(\bx_{jk},\by_{jk})$. Another word we shuffle the first and second elements of pairs in the set $\mathcal{Z}''_{n''}$. 
This means that for each
 $\bz_{ik}=(\bx_{ik},\by_{ik})\in\mathcal{Z}''_{n''}$ given the first element $\bx_{ik}$ the second element $\by_{ik}$ is replaced by a randomly selected $y\in\{y_{jk}\}_{j=1}^{n''}$. This results in a shuffling of the binary relation relating $y_{ik}$ in $y_{jk}$.  
The estimator of $I_p(\BX;\BY)$ is derived based on the Friedman-Rafsky (FR) multivariate runs test statistic \cite{FR} on the concatenated data set , $\mathcal{Z}'_{n'}\cup \widetilde{\mathcal{Z}}_{n''}$. The FR test statistic is defined as the number of edges in the MST spanning the concatenated data set that connect a point in $\mathcal{Z}'_{n'}$ to a point in $\widetilde{\mathcal{Z}}_{n''}$. This test statistic is denoted by $\mathfrak{R}_{n',n''}:=\mathfrak{R}_{n',n''}(\mathcal{Z}'_{n'},\widetilde{\mathcal{Z}}_{n''})$. Note that since the MST is unique with probability one (under the assumption that all density functions are Lebesgue continuous) then all inter point distances between nodes are distinct. This estimator converges to $I_p(\BX;\BY)$ almost surely as $n\rightarrow \infty$. 
The procedure is summarized in Algorithm 1.
\begin{algorithm}\label{algorithm1}
 \caption{FR estimator of GMI}
 \begin{algorithmic}[1]
 \renewcommand{\algorithmicrequire}{\textbf{Input:}}
 \renewcommand{\algorithmicensure}{\textbf{Output:}}
 \REQUIRE Data set $\mathcal{Z}_n:=\big\{(\bx_i,\by_i)_{i=1}^n\big\}$
 %$C^U_{X}$, $C^U_{Y}$, $C^L_{XY}$ 
 \vspace{0.2cm}
%  \\ \textit{Initialization} :
  \STATE Find $\tilde{\alpha}$ using arguments in Subsection \ref{subsec:opt.alpha} 
  \vspace{0.2cm}
  \STATE $n'\leftarrow \tilde{\alpha} n$, $n'' \leftarrow (1-\tilde{\alpha}) n$\\
 \vspace{0.2cm}
  \STATE Divide $\mathcal{Z}_n$ into two subsets $\mathcal{Z}'_{n'}$ and $\mathcal{Z}''_{n''}$\\
 \vspace{0.2cm}
  \STATE $\widetilde{\mathcal{Z}}_{n''}\leftarrow \big\{(\bx_{ik},\by_{jk})_{k=1}^{n''}$: shuffle first and second elements of pairs in $\mathcal{Z}''_{n''}\big\}$
  \vspace{0.2cm}
  \STATE $\widehat{\mathcal{Z}}\leftarrow \mathcal{Z}'_{n'}\cup \widetilde{\mathcal{Z}}''_{n''}$\\
  \vspace{0.2cm}
  \STATE Construct MST on $\widehat{\mathcal{Z}}$\\
  \vspace{0.2cm}
  \STATE $\mathfrak{R}_{n',n''}\leftarrow\#$ edges connecting a node in $\mathcal{Z}'_{n'}$ to a node of $\widetilde{\mathcal{Z}}_{n''}$\\
 \vspace{0.2cm}
  \STATE $\widehat{I}_p\leftarrow 1-\mathfrak{R}_{n',n''}\diy\frac{n'+n''}{2n'n''}$
  \vspace{0.2cm}
  \ENSURE  $\widehat{I}_p$, where $p=\tilde{\alpha}$
 \end{algorithmic} 
 \end{algorithm}
\begin{figure}[h]
\centering
  \includegraphics[width=0.7\linewidth]{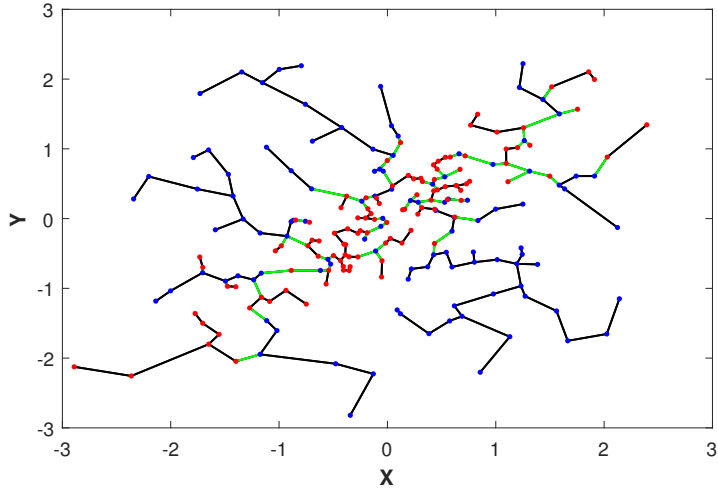}
  \caption{{\small The MST and FR statistic of spanning the merged set of Normal points when $\BX$ and $\BY$ are independent (denoted in blue points) and when $\BX$ and $\BY$ are highly dependent (denoted in red points). The FR test statistic is the number of edges in the MST that connect samples from different color nodes (denoted in green) and it is used to estimate the GMI $I_p$.}}
  \label{fig0}
\end{figure}
Theorem \ref{FR:approximation} shows that the output in Algorithm 1 estimates the GMI with parameter $p={\alpha}$. The proof is provided in Appendix D, Subsection \ref{Proof.FR:approximation}.
\begin{Theorem}\label{FR:approximation}
For given proportionality parameter $\alpha\in (0,1)$, choose $n'$, $n''$ such that $n'+n''=n$ and as $n\rightarrow \infty$, we have $n'/n\rightarrow \alpha$ and $n''/n \rightarrow \beta=1-\alpha$. Then
\beq\label{app:MIP} 1-\mathfrak{R}_{n',n''}(\mathcal{Z}'_{n'},\widetilde{\mathcal{Z}}_{n''})\;\diy\frac{n}{2 n'\;n''} \rightarrow  I_{\alpha}(\BX;\BY), \; \; \;  a.s.\eeq
\end{Theorem}
Note that the asymptotic limit in (\ref{app:MIP}) depends on the proportionality parameter $\alpha$. Later in Subsection \ref{subsec:opt.alpha}, we discuss on choosing an optimal parameter $\tilde{\alpha}$.  In Fig. \ref{fig0}, we show a visualization of the MST constructed over merged independent ($\rho=0$) and highly dependent ($\rho=0.9$) data sets drawn from two dimensional standard Normal distribution with correlation coefficient $\rho$. Notice that the MST connects different colored samples, corresponding to independent and dependent samples, respectively, indicated in green. The total number of green edges is the FR test statistic.

\subsection{Convergence Rates}\label{Con:rates}
\def\ep{\epsilon}
In this subsection we provide the MSE convergence rates in the form of upper bounds on the bias and the variance. This MSE bound is given in terms of the sample size $n$, the dimension $d$, and the proportionality parameter $\alpha$. Deriving convergence rates for mutual information estimators has been of interest in information theory and machine learning \cite{PPS}, \cite{MSH2017}. The rates are typically derived in terms of a smoothness condition on the densities, such as the H\"{o}lder condition \cite{Wolfgang1990}. Here we 
assume $f_{X}$, $f_{Y}$ and $f_{XY}$ with support sets $\bbS_X$, $\bbS_Y$, and $\bbS_{XY}:=\bbS_X\times\bbS_Y$, respectively, are smooth in the sense that they belong to strong H\"{o}lder continuous classes of densities $\Sigma^s_d(\eta,$ K), $0<\eta\leq 1$ \cite{Lo}, \cite{Wolfgang1990}:
\begin{Definition}\label{def:strong.Holder}
(Strong H\"{o}lder class): Let $\mathcal{X}\subset \bbR^d$ be a compact space. The strong H\"{o}lder class $\Sigma_d^s(\eta,K)$, with $\eta$-H\"{o}lder parameter and constant $K$, of functions on $L_d$-norm, consists of the functions $g$ that satisfy 
\beq\label{SHC}\begin{array}{l}\Big\{g: \big\|g(\bz)-p_{\bx}^{\lfloor \eta \rfloor}(\bz)\big\|_d\leq K\; g(\bx)\; \big\|\bx-\bz\big\|_d^\eta,\;\; \bx,\;\bz\in \mathcal{X}\Big\}, \end{array}\eeq
where $p_{\bx}^k(\bz)$ is the Taylor polynomial (multinomial) of $g$ of order $k$ expanded about the point $\bx$ and $\lfloor\eta\rfloor$ is defined as the greatest integer strictly less than $\eta$. Note that  for the standard H\"{o}lder class the factor $g(\bx)$ on the RHS of (\ref{SHC}) is omitted. 
\end{Definition}
To find the optimal parameter $\alpha$ we require explicit expressions for the bias and variance. Bounds are provided in Appendix E, Subsection \ref{Proof:th.Optimized:Bias}. To obtain such expressions, we require several assumptions on the absolutely continuous densities $f_X$, $f_Y$, $f_{XY}$ and support sets $\bbS_X$, $\bbS_Y$, $\bbS_{XY}$:
\vspace{0.5cm}
\begin{itemize} [noitemsep,nolistsep,itemindent=10pt]
 \setlength\itemsep{0.7em}
\item[({\bf A}.1)] Each of the densities belong to $\diy \Sigma^s_d(\eta,K)$ class in their support sets with smoothness parameters $\eta$ and smoothness constants $K$.
%$K=K_{X}$, $K_{Y}$, and $K_{XY}$, respectively.  
\bigskip
\item[({\bf A}.2)] The volumes  of the support sets are finite, i.e. $0<\bbV(\bbS_{X})<\infty,\;\;\;0<\bbV(\bbS_{Y})<\infty$. %$$0<\bbV(\bbS_{X})<\infty,\;\;\;0<\bbV(\bbS_{Y})<\infty,\;\;\; 0<\bbV(\bbS_{XY})<\infty.$$
\bigskip
\item[({\bf A}.3)] All densities are bounded on their support sets, i.e. there exist two sets of constants $C^L_{X},C^L_{Y},C^L_{XY}$ and  $C^U_X, C^U_Y,C^U_{XY}$ such that $0<C^L_{X}\leq f_{X}\leq C^U_{X}<\infty$, $0<C^L_{Y}\leq f_{Y}\leq C^U_{Y}<\infty$ and $0<C^L_{XY}\leq f_{XY}\leq C^U_{XY}<\infty$.  
\end{itemize}
\vspace{1cm}
{\bf Remarks:}
\begin{itemize}
    \item [1.] The assumption ({\bf A}.3) implies that the strong H\"{o}lder class, ({\bf A}.1), is a subset of standard  H\"{o}lder class:
    \begin{equation}\label{HC}\Big\{g: \big\|g(\bz)-p_{\bx}^{\lfloor \eta \rfloor}(\bz)\big\|_d\leq K'\; \big\|\bx-\bz\big\|_d^\eta,\;\; \bx,\;\bz\in \mathcal{X}\Big\}. \end{equation}
    \item [2.] If in ({\bf A}.3) $C^L_{X},C^L_{Y},C^L_{XY}$ and  $C^U_X, C^U_Y,C^U_{XY}$ are equal i.e. the density functions are uniform, $g(\bx)=K''$, and if $K'=K.K''$ then two classes (\ref{SHC}) and (\ref{HC}) are equal.
    \item [3.] By using the Strong H\"{o}lder class (\ref{SHC}) our proposed bias bound in Theorem \ref{th.Optimized:Bias} below, becomes tighter. Another bound can be derived for the standard H\"{o}lder class (\ref{HC}) but the proposed bias bound becomes weaker, and so in the result for bounding convergence rate.
\end{itemize}

The following theorems on the bias and variance follow under assumptions ({\bf A}.1), ({\bf A}.2), and ({\bf A}.3):  
\begin{Theorem}\label{th.Optimized:Bias}
Under two assumptions ({\bf A}.1) and ({\bf A}.3), for given $\alpha\in(0,1)$, $\beta=1-\alpha$, the  bias of the $\mathfrak{R}_{n',n''}$ estimator for $d\geq 2$, $0<\eta\leq 1$ satisfies

\begin{equation}\label{Optimized:Bias}\begin{array}{l}\diy \bigg|\frac{\bbE\big[\mathfrak{R}_{n',n''}\big]}{n}-2\alpha \beta\iint \frac{f_{\BX\BY}(\bx,\by) f_{\BX}(\bx) f_{\bY}(\by)}{\alpha f_{\BX\BY}(\bx,\by)+\beta f_{\BX}(\bx)f_{\BY}(\by)}\;\rd\bx\rd\by \bigg|\\
\\
\qquad\quad\leq O\bigg(\max \Big\{n^{-\eta^2\big/(d(1+\eta))},\; (\beta n)^{-\eta/(1+\eta)},\; c_d 2^d n^{-1}\Big\}\bigg), 
\end{array}\end{equation}
where $c_d$ is a constant depending only on $d$.
\end{Theorem}
\vspace{-0.1cm}
Note that according to Theorem 13 in \cite{Robins1994}, the constant $c_d$ is lower bounded by $\Omega\left(\sqrt{d} 2^{n(1-H(\gamma))}\right)$, $\gamma=2^{-d}$ and $H(\gamma)$ is the binary entropy i.e.
$$H(\gamma)=-\gamma\log \gamma -(1-\gamma)\log (1-\gamma).$$

A proof of Theorem \ref{th.Optimized:Bias} is given in Appendix E. The next theorem gives an upper bound on the variance of the FR estimator $\mathfrak{R}_{n',n''}$. The proof of the variance result requires a different approach than the bias bound. The proof uses the Efron-Stein inequality \cite{ES} and is similar to arguments in \cite[Appendix C]{YNMH}, and is omitted. In Theorem \ref{variance:main} we assume that the densities $f_X$, $f_Y$, and $f_{XY}$ are absolutely continuous and bounded. Note that we can obtain a weaker bound than (\ref{Optimized:Bias}) when we relax ({\bf A}.1) to the standard H\"{o}lder class with a constant coefficient $K'$ instead of $K g(\bx)$. 
%in their support sets with smoothness parameters $\eta$ and smoothness constants $K'$ but then the derived bias bound doesn't become as tight as the bound given in (\ref{Optimized:Bias}).
\begin{Theorem}\label{variance:main} Given $\alpha\in(0,1)$, the variance of the estimator $\mathfrak{R}_{n',n''}(\mathcal{Z}'_{n'},\widetilde{\mathcal{Z}}_{n''})$ is bounded by
\beq \label{var.bound} Var\left(\diy\frac{\mathfrak{R}_{n',n''}}
%(\BZ'_{n'},\widetilde{\BZ}_{n''})}
{n}\right)
\leq \diy\frac{{(1-\alpha)\; c_d}}{n}, \;\;\; \alpha=n'/n,\eeq
where $c_d$ is a constant depending only on the dimension $d$.
\end{Theorem}
%-----------------------------
\subsection{Minimax Parameter $\alpha$}\label{subsec:opt.alpha}
A question in Algorithm 1 is the existance of a best proportion $\alpha$ that partitions sample. In this subsection we obtain on an upper bound of the MSE rate and characterize the parameter $\alpha$ that minimizes the upper bound. This optimal $\alpha$ depends on maximize densities, however.
%This discussion will help up to divide the sample into optimal proportions such that the MSE rate bound is minimax.
% another word the estimation is more accurate. We establish this theory so that it can be applied to a wider variety of estimation problems. 

Recall assumptions ({\bf A}.1), ({\bf A}.2), and ({\bf A}.3) in Subsection \ref{Con:rates}. The constant $\alpha$ can be chosen to ensure that the MSE converges rate obtained from the bias and variance rates  (\ref{Optimized:Bias}) and (\ref{variance:main}) by selecting $\alpha$ to minimize the maximum MSE, where the maximum is taken over the space of H\"{o}lder smooth joint densities $f_{XY}$.\\
\\
Throughout this subsection we use the following notations:
\begin{itemize}
    \item $\epsilon_{XY}:=f_{XY}\big/f_{X}f_{Y}$,
    \vspace{0.3cm}
    \item $C^L_{\epsilon}:=C^L_{XY}\big/C^U_{X}C^U_{Y}$ and $C^U_{\epsilon}:=C^U_{XY}\big/C^L_{X}C^L_{Y}$, 
    \vspace{0.3cm}
    \item $C_n:=C^L_{XY}\; n/2$,
    \vspace{0.3cm}
    \item $\alpha^L_0:=\diy\frac{2}{C_n}$ and $\alpha^U_0:=\min\left\{\diy\frac{1}{4}, \diy\frac{1+1/C_n}{4+2C^U_\ep}, 1-n^{\eta/d-1}\right\}$, where $\eta$ is the smoothness parameter,
    \vspace{0.3cm}
    \item $l_n:=\big\lfloor n^{\eta/(d^2(1+\eta))}\big\rfloor$.
\end{itemize}
Now define $\widetilde{{G}}^{\alpha,\beta}_{\epsilon_{XY},n}(\bx,\by)$ by 
\begin{equation}
\diy\frac{(\epsilon_{XY}(\bx,\by)+1/(\beta C_n))(1+\epsilon_{XY}(\bx\by)+1/(\beta C_n))}{(\alpha+\beta \epsilon_{XY}(\bx,\by))^2},\;\;\; \beta=1-\alpha.
\end{equation}
Consider the following optimization problem: 
\def\bt{\mathbf{t}}
\begin{equation}\label{opt:problem1}\begin{aligned}
& \underset{\alpha}{\text{min}}\;\;\underset{\epsilon_{XY}}{\text{max}}
& & \widetilde{\Delta}(\alpha,\epsilon_{XY})+c_d(1-\alpha)\;n^{-1}\\
& \text{subject to}
& & C^L_{\epsilon}\leq  \epsilon_{XY}\leq C^U_{\epsilon},\\
& & &\alpha^L_0\leq \alpha\leq \alpha^U_0,
\end{aligned}\end{equation}
where  
\begin{equation}\label{def:var1} 
\diy\widetilde{\Delta}(\alpha,\epsilon_{XY}):=\diy D(n,l_n,d,\eta)+\overline{D}(n,l_n,d)C^U_{XY}\diy\iint_{\bbS_{XY}} \widetilde{{G}}^{\alpha,\beta}_{\epsilon_{XY},n}(\bx,\by)\;\rd\bx\rd\by,
\end{equation}
and
%\begin{equation}
%D(n,l_n,d,\eta)=
%\end{equation}
%\begin{equation}
%\overline{D}(n,l_n,d)=
%\end{equation}
\beq \label{Def.D}\begin{array}{l} D(n,l_n,d,\eta)=\diy c_2l_n^dn^{-1}+c_d2^dn^{-1}+c'l_n^d n^{-\eta/d}+\diy c l_n^d n^{-1/d}+2c_1l_n^{d-1}n^{1/d-1},\end{array}\eeq
\begin{equation}\label{tilde-D}\begin{array}{l} \widetilde{D}(n,l_n,d) 
=\diy 2+n^{-1}2c''\sum\limits_{i=1}^M l_n\; l_n^d a_i^{-1}+n^{-3/2}2 c'_1\sum\limits_{i=1}^M l_n\; l_n^{d/2}\sqrt{b_i}{a_i^2}\\
\qquad+n^{-1}\diy\sum\limits_{i=1}^M 2 n^{-3/2}l_n^{-d/2}\diy\frac{\sqrt{b_i}}{a_i^2}\big(na_il_n^d+n^2a_i^2\big)^{1/2}\big(nb_il_n^d+n^2b_i^2\big)^{1/2}.\end{array}\end{equation}
Note that in (\ref{Def.D}), $c,c',c_1,c_2$ are constants, and $c_d$ only depends on the dimension $d$. Also, in (\ref{tilde-D}), $a_i$ and $b_i$ are constants. Let $\epsilon_{XY}^*$ be the optimal $\ep_{XY}$ i.e. $\ep_{XY}^*$ be the solution of the optimization problem (\ref{opt:problem1}). Set  
%See Appendix F.\\

%{\textit{More Notations}:} Define 
%$$\theta^U(\alpha):=\diy\frac{1-4\alpha+1/C_n}{2\alpha},$$
%$$\theta^U_\epsilon(\alpha):=\min\left\{C^U_\epsilon, \theta^U(\alpha)\right\},\;\;\;\; \theta^L_\epsilon(\alpha):=C^L_\epsilon,$$
%$$\alpha^L_0:=\diy\frac{2}{C_n},\;\;\; \alpha^U_0:=\min\left\{\diy\frac{1}{4},1-n^{\eta/d-1}\right\}.$$
%$$\diy\frac{2}{C_n}\leq \alpha\leq \min\big\{\diy\frac{1}{4}, \diy\frac{1+1/C_n}{4+2C^U_\ep}, 1-n^{\eta/d-1}\big\}.$$

\begin{equation}\label{def.Xi}\begin{array}{l} \Xi(\alpha):=\diy\frac{\rd}{\rd \alpha} \Big(\widetilde{\Delta}(\alpha,\epsilon^*_{XY})+c_d(1-\alpha)\;n^{-1}\Big),\\
%=\overline{D}(n,l_n,d)\;C^U_{XY}\bbV(\bbS_{XY})\;\diy\frac{\rd}{\rd \alpha}\widetilde{{G}}^{\alpha,\beta}_{\epsilon^*_{XY},n}-c_d\; n^{-1}.
\end{array}\end{equation}
%Such that $\epsilon^*_{XY}$ depends only on $n$ and $\alpha$.
%Consider the optimization problem (\ref{opt:problem1}). 
such that $\diy\widetilde{\Delta}(\alpha,\epsilon^*_{XY})$ is (\ref{def:var1}) when $\epsilon_{XY}=\epsilon^*_{XY}$. For $\alpha\in [\alpha^L_0,\alpha^U_0$], the optimal choice of $\epsilon_{XY}$ in terms of maximizing the MSE is $\epsilon_{XY}^*=C^U_\epsilon$ and the saddle point for the parameter $\alpha$, denoted by $\tilde{\alpha}$, is given as follows:
\begin{center}
\begin{itemize}
\item  $\tilde{\alpha}=\alpha_0^U$, if $\Xi(\alpha_0^U)<0$. 
\item $\tilde{\alpha}=\alpha_0^L$, if $\Xi(\alpha_0^L)>0$. 
\item $\tilde{\alpha}=\Xi^{-1}(0)$, if $\alpha_0^L\leq \Xi^{-1}(0)\leq \alpha_0^U$.
\end{itemize}
\end{center}

Further details are given in Appendix F. 
%Overall, the optimized solution for $\alpha$ leads us to establish the proposed algorithm, i.e. Algorithm 1, for the FR estimator of the GMI measure, $I_p$, where $p=\tilde{\alpha}$. Below in Section 3 (Simulation Study), we perform sets of experiments for various sample sizes $n$, dimensions $d$, and parameters $\alpha$. We numerically observe that in the experiments when $\alpha=0.5$ the MSE is less than when $\alpha=0.2\;\;\hbox{or}\;\;0.8$, see Table 1. This shows that the optimization problem in (\ref{opt:problem1}) is restricted to $\alpha\in[\alpha_0^L,\alpha_0^U]$, where $\alpha_0^U$ is not larger than $\diy\frac{1}{4}$. That is, for the parameter $\alpha$ in the complement of $[\alpha_0^L,\alpha_0^U]$ the solution given above is not necessarily guaranteed.  

\section{Simulation Study}\label{experiments}
In this section numerical simulations are presented that illustrate the theory in Section \ref{sec:HP-MI}. We perform multiple experiments to demonstrate the utility of the proposed direct estimator of the HP-divergence in terms of the dimension $d$ and the sample size $n$. Our proposed MST-based estimator of the GMI is compared to other plug-in GMI estimators, in particular the standard KDE estimator of \cite{MSH2017}, where the convergence rates of Theorem \ref{th.Optimized:Bias} and \ref{variance:main} are validated. We use multivariate Normal simulated data in the experiments. In this section, we also discuss the choice of the proportionality parameter $\alpha$ and compare runtime of our proposed FR estimator approach with KDE method.

Here we perform four sets of experiments to illustrate the estimator and the theory derived above. For the first set of experiments the MSE of the FR estimator proposed in Algorithm 1 is shown in  Fig. 2-(left). The samples were drawn from $d$-dimensional Normal distribution, with various sample sizes and dimensions $d=6,10,12$. We selected the proportionality parameter $\alpha=0.3$ and computed the MSE in terms of the sample size $n$. We show the log-log plot of MSE when $n$ varies in $[100,1500]$. Note that the empirically optimal proportion $\alpha$ depends on $n$, so to avoid the computational complexity we fixed $\alpha$ for this experiment. The experimental result shown in Fig. 2-(left) validates the theoretical MSE growth rates derived from (\ref{Optimized:Bias}) and (\ref{var.bound}), i.e., decreasing sublinearly in $n$ and increasing exponentially in $d$. 
\begin{figure}[h]
%\centering
  \begin{subfigure}[b]{0.4\textwidth}
    \includegraphics[width=1.2\textwidth]{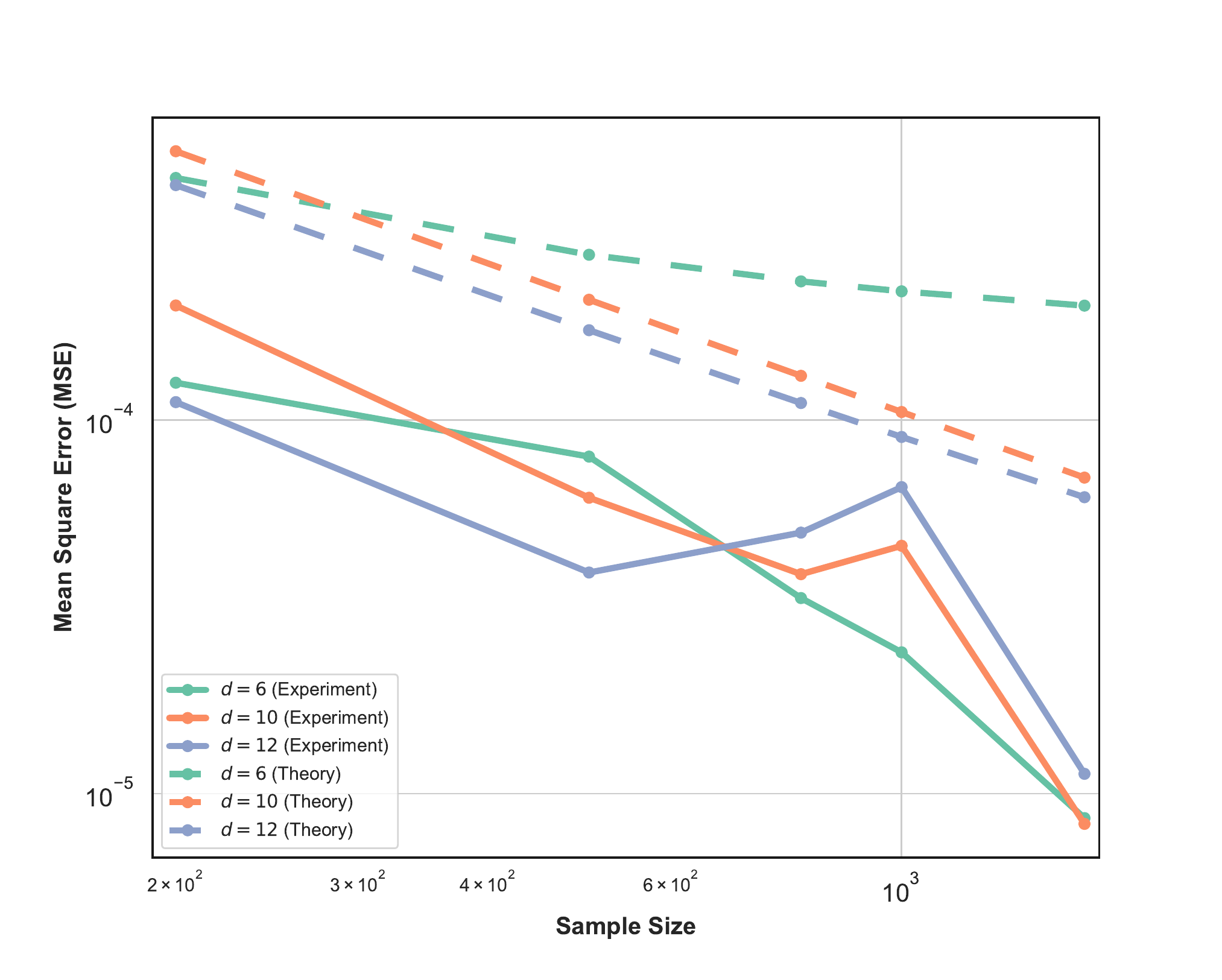}
    %\caption{{\small Theoretical and experimental log(MSE) of the FR estimator of GMI as a function of sample size $n$ for $d=6, 10, 12$.}}
   % \label{fig:1}
  \end{subfigure}
  \hspace{1.2cm}
  \begin{subfigure}[b]{0.4\textwidth}
    \includegraphics[width=1.2\textwidth]{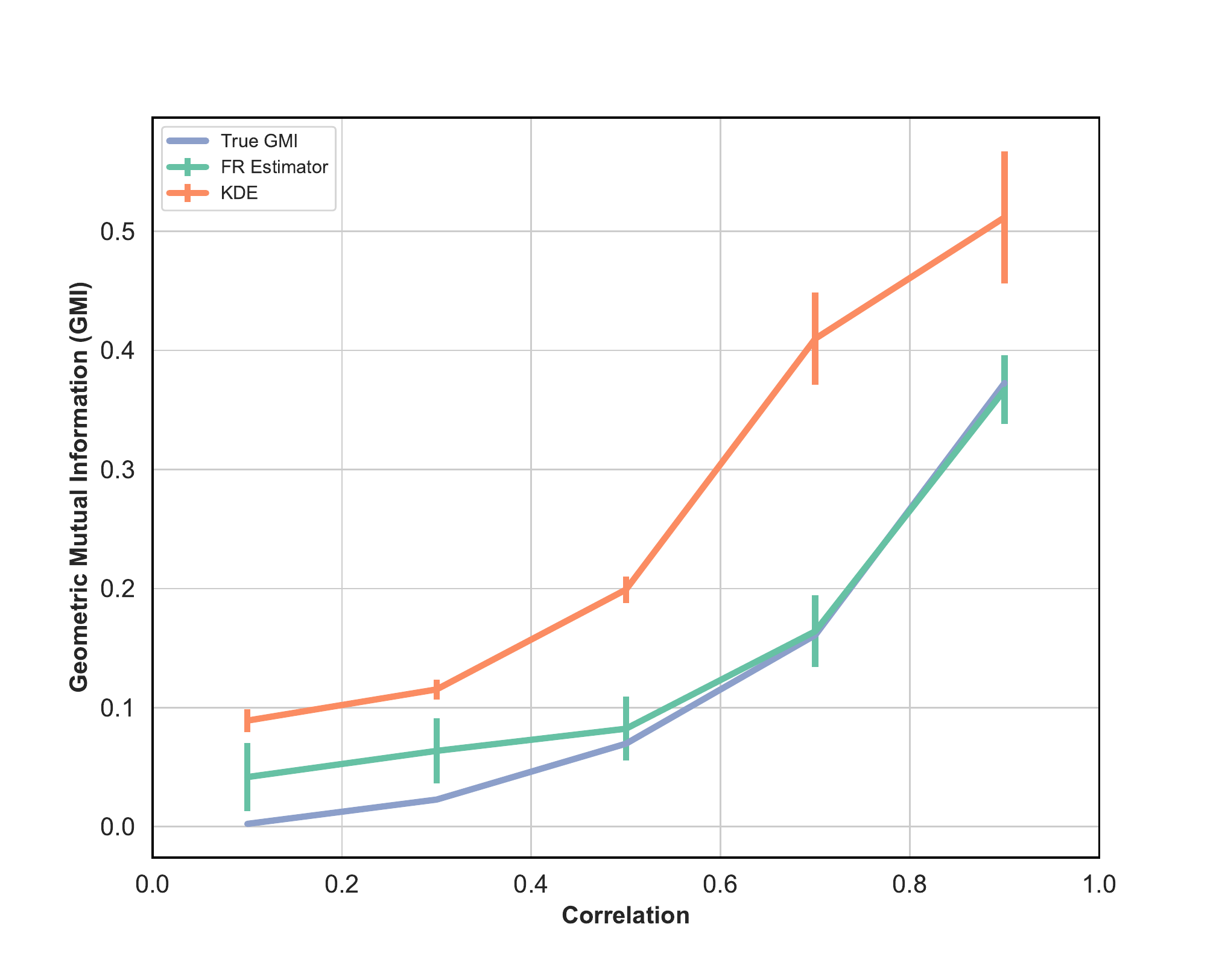}
  %  \caption{{\small The GMI estimator using two approaches FR statistics and ensemble KDE along with True GMI for $\alpha=0.6$.}}
   % \label{fig:2}
  \end{subfigure}

  \caption{(left) Log–log plot of theoretical and experimental MSE of the FR estimator of the GMI as a function of sample size $n$ for $d=6, 10, 12$ and fixed smoothness parameter $\eta$. (right) The GMI estimator using two approaches, FR test statistic and KDE method along with True GMI. In this experiment, we generated data from the two dimensional Normal distribution with zero mean and covariance matrix (\ref{cor.matrix}) for various $\rho\in[0.1,0.9]$.}
    \label{fig:2}
\end{figure}

In Fig. 2-(right), we compare our estimator with the true GMI and the standard KDE \cite{MSH2017}. For the KDE approach, we estimated the joint and marginal densities and then plugged them into the proposed expression (3) in \cite{MSH2017}. The bandwidth used for the the KDE plug-in estimator was selected by setting $h= n^{-1/(d+1)}$ to minimize the MSE of the plug in estimator. We generated data from the two dimensional Normal distribution with zero mean and covariance matrix 
\begin{equation}\label{cor.matrix}
\left(\begin{array}{cc} 1 & \rho\\ \rho & 1 \end{array}\right).
\end{equation}
We varied $\rho$ in range $[0.1,0.9]$ and computed the proposed estimated GMI and the KDE. We used the Monte Carlo method to approximate the integral (\ref{EQ:MIP}) and compute the GMI measure (labeled True GMI). We see that as $\rho$ increases the estimated GMI using the FR test statistic outperforms the KDE approach (FR curve approaches to True curve and KDE curve diverges from True curve). In this set of experiments $\alpha=0.6$. 
\begin{figure}[h]
%\centering
  \begin{subfigure}[b]{0.4\textwidth}
    \includegraphics[width=1.2\textwidth]{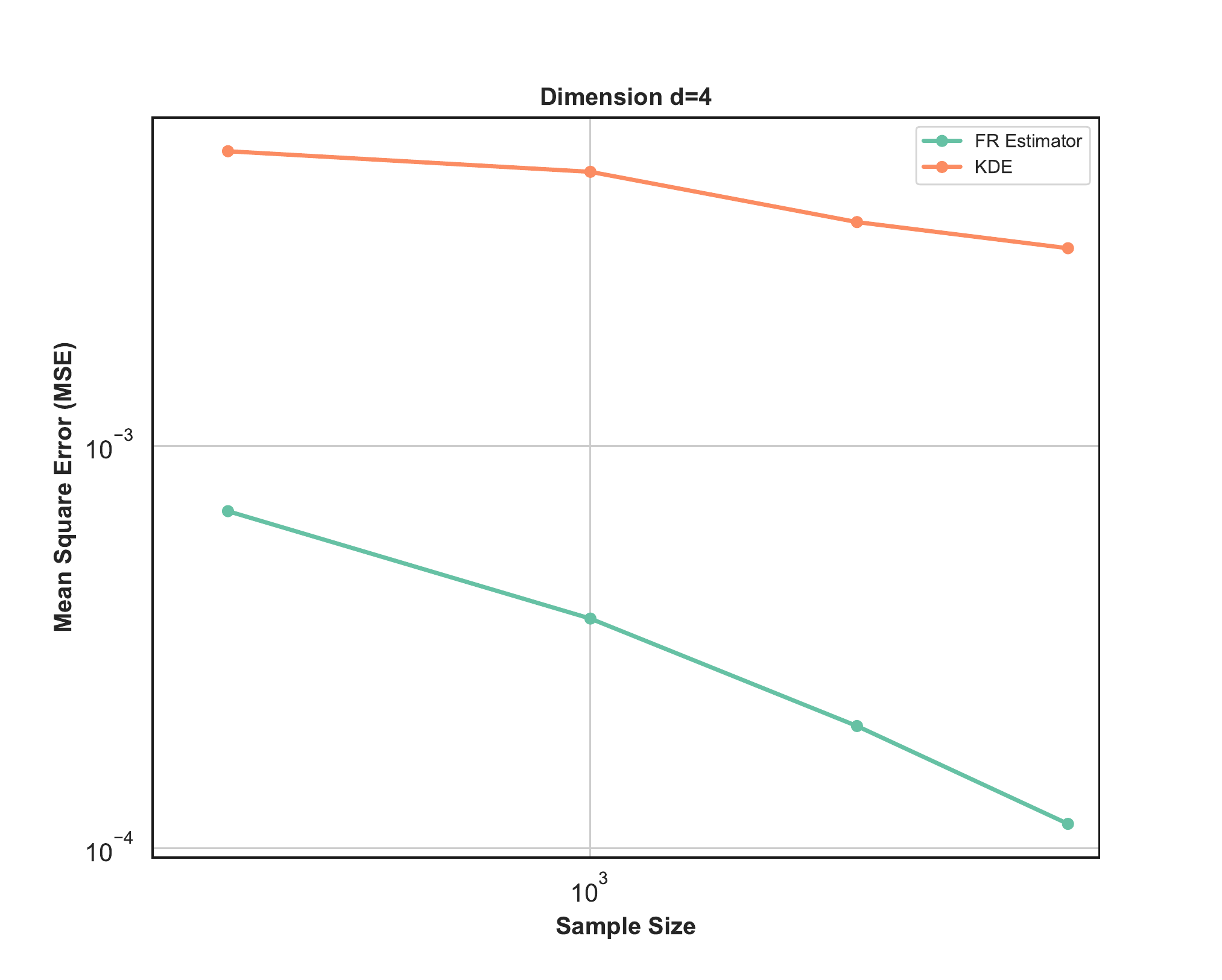}
    %\caption{{\small ..}}
    \label{fig:1}
  \end{subfigure}
  \hspace{1.2cm}
  \begin{subfigure}[b]{0.4\textwidth}
    \includegraphics[width=1.2\textwidth]{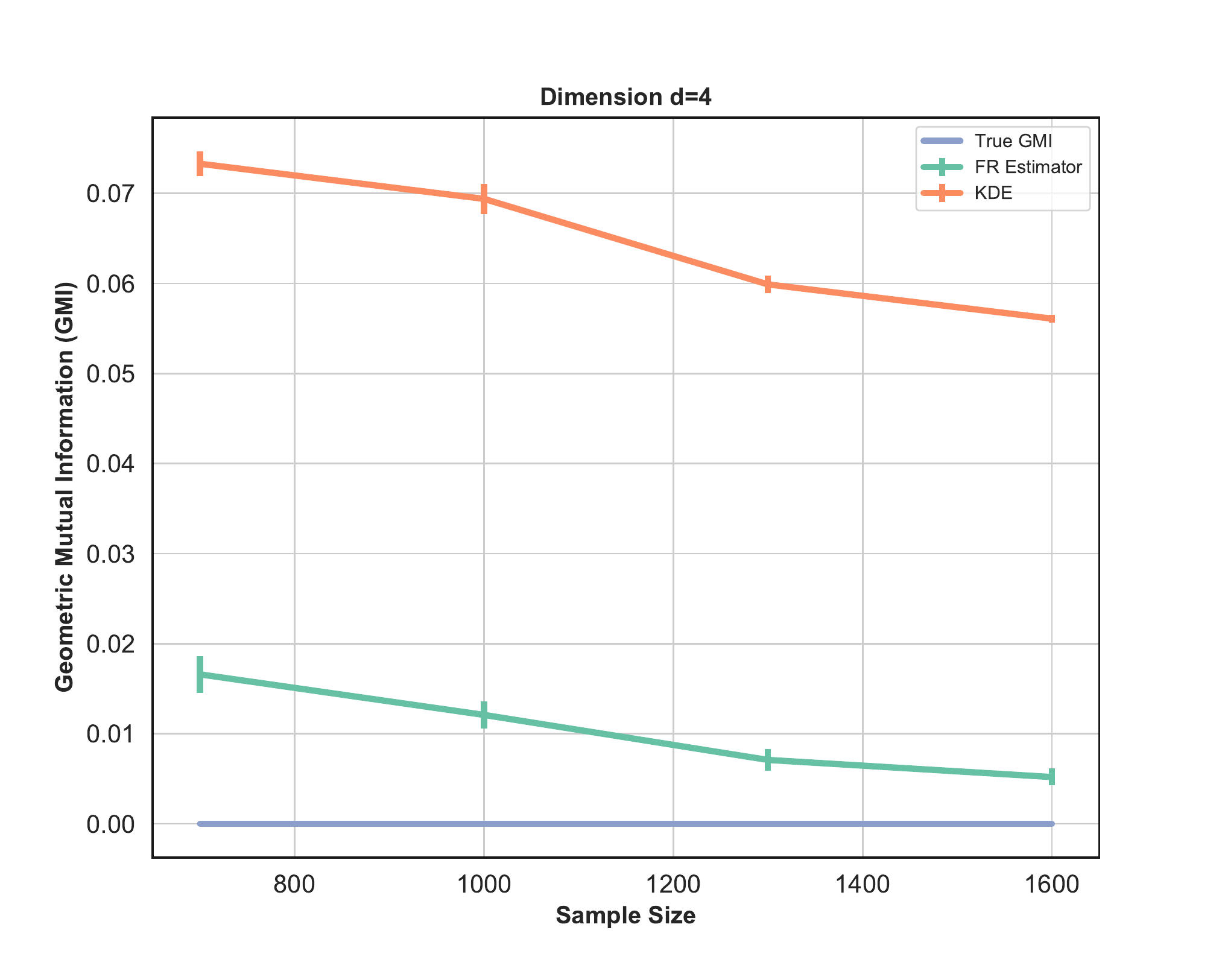}
    %\caption{{\small }}
    \label{fig:2}
  \end{subfigure}\\
  \begin{subfigure}[b]{0.4\textwidth}
    \includegraphics[width=1.2\textwidth]{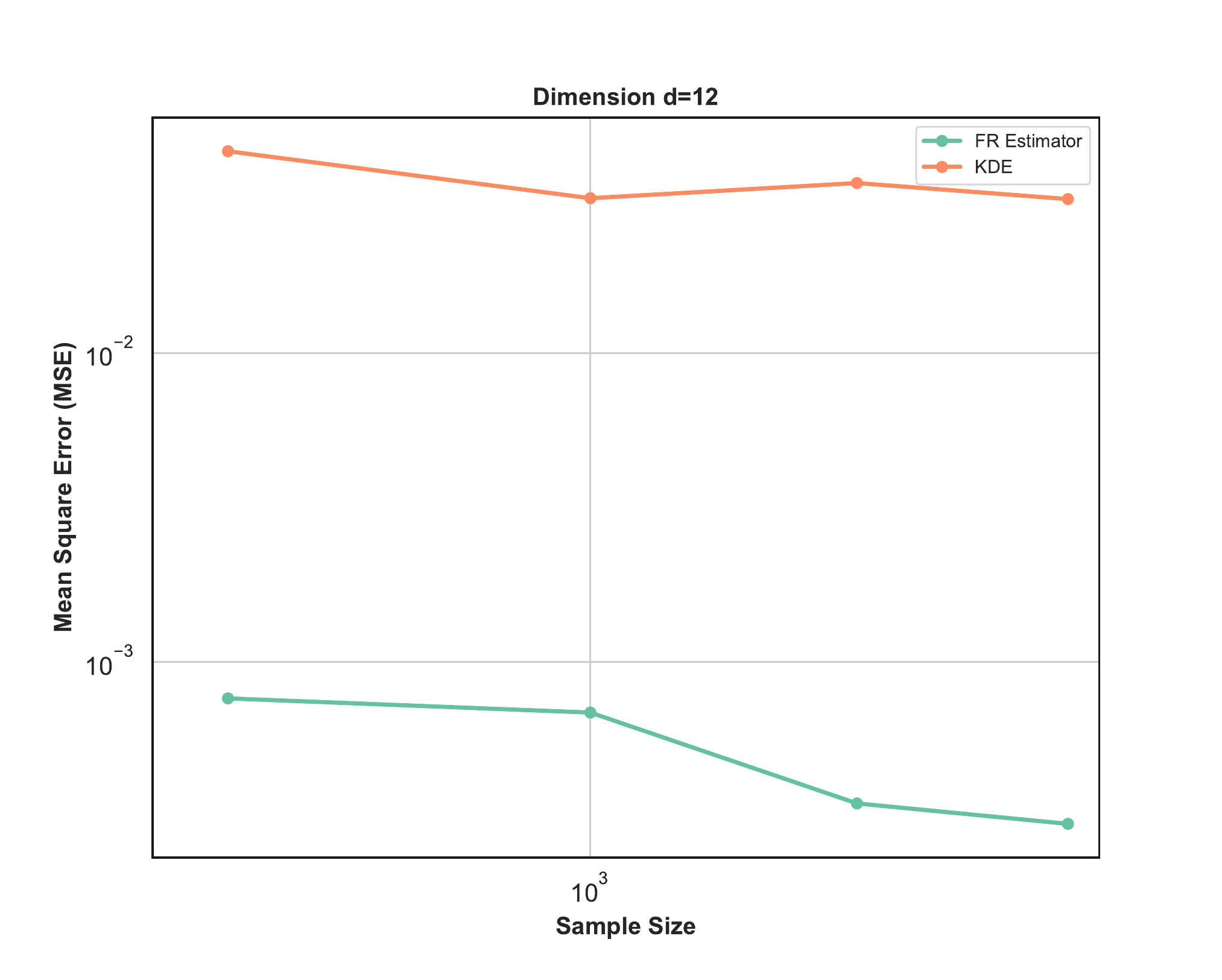}
    %\caption{{\small ..}}
  \end{subfigure}
  \hspace{1.2cm}
  \begin{subfigure}[b]{0.4\textwidth}
    \includegraphics[width=1.2\textwidth]{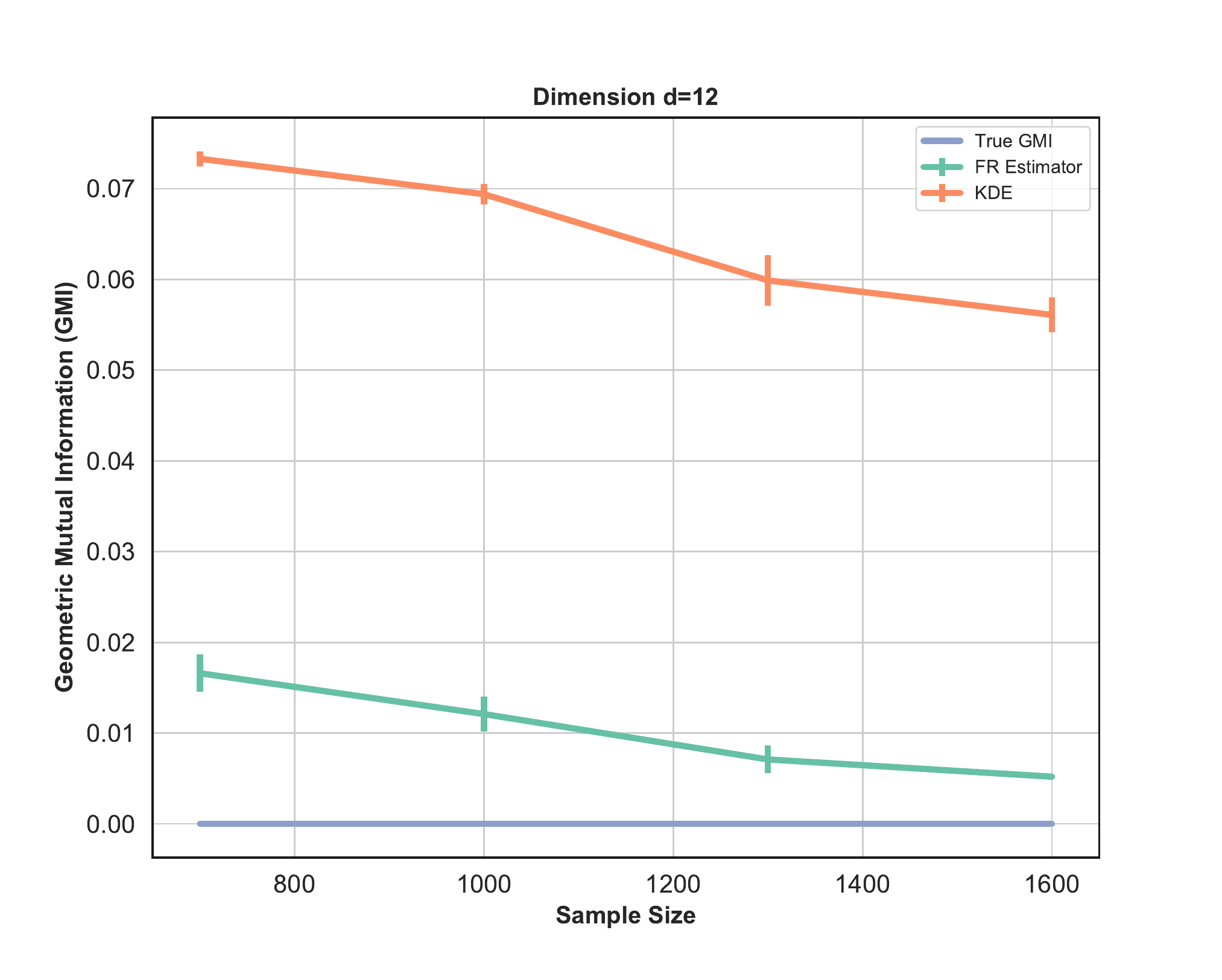}
    %\caption{{\small }}
  \end{subfigure}
  \label{FRvsKDE}
  \caption{MSE log-log plots as a function of sample size $n$ (left) for the proposed Friedman-Rafsky estimator ("Estimated GMI") and the standard KDE plug-in estimator ("KDE"). The two right plots each corresponds to the geometric mutual information with the dimension $d=4$ (top) and $d=12$ (bottom). In both cases the proportionality parameter $\alpha$ is $0.5$.  The FR estimator in both plots for sample size $n$ in $[700,1600]$ outperforms the standard kernel plug-in estimator, especially for larger dimensions. }
\end{figure}

Fig. 3 again compares the FR estimator with the standard KDE estimator. In this setting, we draw samples from the multivariate standard Normal distribution with dimensions $d=4$ and $d=12$. In both cases the proportionality parameter $\alpha=0.5$. The left plots in Fig. 3 show the MSE (100 trials) of the GMI estimator implemented with an KDE estimator (with bandwidth as in Fig. 2 i.e. $h=n^{-1/(d+1)}$) for dimensions $d=4,12$ and various sample sizes. For all dimensions and sample sizes the FR estimator also outperforms the plug-in KDE estimator based on the estimated log-log MSE slope given in Fig. 3 (left plots). The right plots in Fig. 3 show the geometric mutual information estimated by KED and FR approaches. The iteration in this experiment is 100 and error bars are standard deviations. We observe that for higher dimension $d=12$ and larger sample size $n$, the KDE estimator approaches to the true GMI slower than the FR estimator. This reflects the power of graph-based (direct) estimators, and in particular our proposed FR estimator.
\vspace{0.2cm}

\begin{figure}[h]
%\centering
 \begin{subfigure}[b]{0.4\textwidth}
    \includegraphics[width=1.2\textwidth]{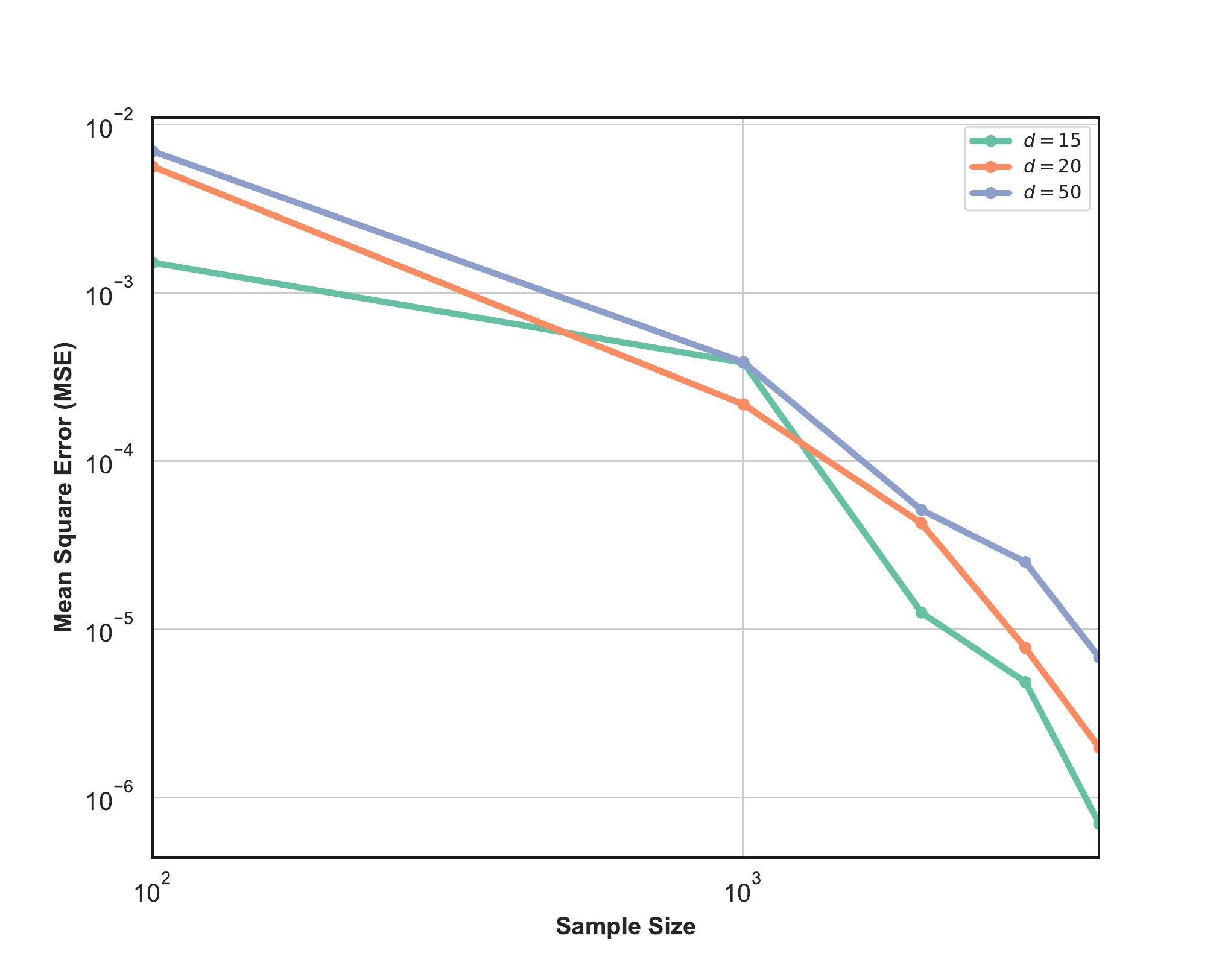}
    \label{fig:1}
  \end{subfigure}
  \hspace{0.8cm}
  \begin{subfigure}[b]{0.425\textwidth}
     \includegraphics[width=1.35\textwidth]{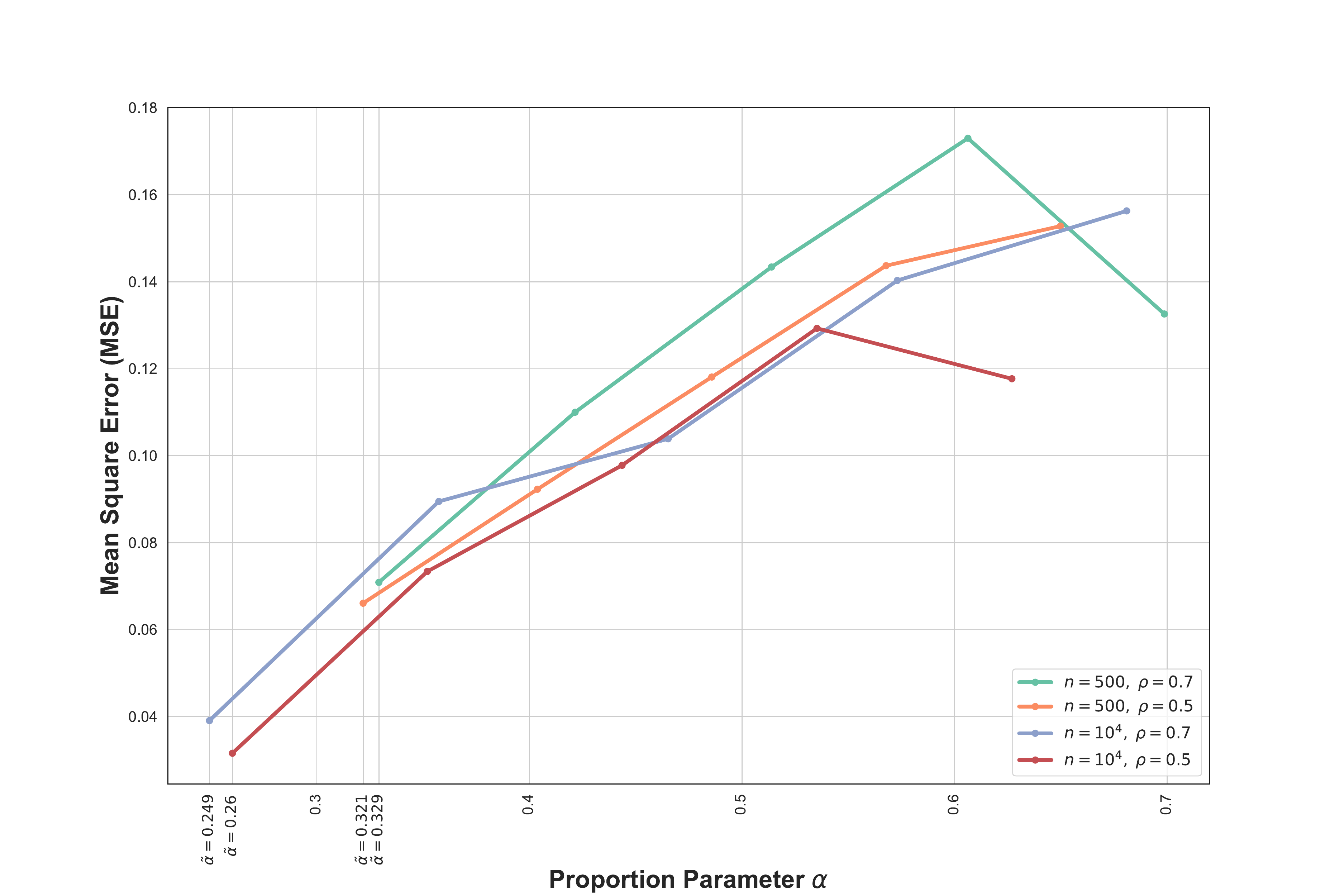}
    %\caption{{\small }}
    \label{fig:2}
  \end{subfigure}\\
  \label{GMIvsN}
  \caption{MSE log-log plots as a function of sample size $n$ for the proposed FR estimator. We compare the MSE of our proposed FR estimator for various dimensions $d=15,20,50$ (left). As $d$ increases, the blue curve takes larger values than green and orange curves i.e. MSE increases as $d$ grows. However, this is more evidential for large sample size $n$. The second experiment (right) focuses on optimal proportion $\alpha$ for $n=500, 10^4$ and $\rho=0.7,0.5$. $\tilde{\alpha}$ is the optimal $\alpha$ for $\alpha\in[\alpha_0^L,\alpha_0^U]$.}
\end{figure}
The comparison between MSEs for various dimension $d$ is shown in Fig. 4 (left). This experiment highlights higher dimension's impact on our proposed FR estimator for the GMI measure. As expected, for larger sample size $n$, MSE decreases while for higher dimension it increases. In this setting, we have generated samples from standard Normal distribution with size $n\in[10^2, 4\times10^3]$ and $\alpha=0.5$. From Fig. 4 (left) we observe that for larger sample size, MSE curves are ordered based on their corresponding dimensions. Fig. 4 (right) illustrates the MSE vs proportion parameter $\alpha$ when $n=500,10^4$ samples are generated from Normal distribution with $\rho=0.7,0.5$. First, following subsection 2.4, we compute the bound $[\alpha_0^L, \alpha_0^U]$ and then derive optimal $\alpha$ in this range. Therefore, each experiment with different sample size and $\rho$ provides different range $[\alpha_0^L, \alpha_0^U]$. We observe that the MSE is not necessarily a monotonic function in $\alpha$ and its behavior strongly depends on sample size $n$, $d$, and density functions' bounds. This is studied extensively in Appendix F. In this set of experiments $\Xi(\alpha_0^L)>0$, therefore following the results in subsection 2.4, we have $\tilde{\alpha}=\alpha_0^L$. The optimal $\alpha$ is indicated in the Fig. 4 (right). 
\vspace{0.2cm}

Next, we consider three scenarios to analyze parameter $\alpha$. In these scenarios the lower bounds $C^L_X,\;C^L_Y,\;C^L_{XY}$ and upper bounds $C^U_X,\;C^U_Y,\;C^U_{XY}$ are unknown, therefore results in Section \ref{subsec:opt.alpha} are not applicable. In these set of experiments we varied $\alpha$ in the range $(0,1)$ to divide our original sample.
%and experimentally detect the optimal proportion to divide our original sample. 
We generated sample from the multivariate standard Normal distribution in all three scenarios (all features are independent). Therefore the true GMI is zero and in all scenarios the GMI column is compared with zero. In each scenario we fixed dimension $d$ and sample size $n$ and varied $\alpha=0.2, 0.5,0.8$. The dimension and sample size in Scenarios 1,2, and 3 are $d=6,8,10$ and $n=1000,1500,2000$, respectively. In Table 1 the last column ($\alpha$) stars the parameter $\alpha\in\{0.2,0.5,0.8\}$ with the minimum MSE and GMI $(I_\alpha)$ in each scenario. Table 1 shows that in these sets of experiments when $\alpha=0.5$, the GMI estimator has less MSE (i.e. is more accurate) than when $\alpha=0.2$ or $\alpha=0.8$. This experimentally demonstrates that if we split our training data, the proposed Algorithm 1 performs better than $\alpha=0.2\;\hbox{or}\;0.8$ to estimate the GMI measure. 
\begin {table}[h]
\begin{center}
\begin{tabular}{ |c||c|c|c|c|c|  }
 \hline
 \multicolumn{6}{|c|}{Overview table for different $d$, $n$, and $\alpha$} \\
 \hline
Experiments & Dimension ($d$)  & Sample size ($n$) & GMI ($I_\alpha$)& MSE ($\times 10^{-4}$)& Parameter ($\alpha)$ \\
 \hline
Scenario 1-1 & 6  & 1000    & 0.0229 &   12&  0.2\\
Scenario 1-2 & 6  & 1000    & 0.0143 &   4.7944&  0.5*\\
Scenario 1-3 & 6  & 1000    & 0.0176 &   6.3867&  0.8\\
Scenario 2-1 & 8  & 1500  &   0.0246   & 11 &  0.2\\
Scenario 2-2 & 8  & 1500  &  0.0074   & 1.6053&  0.5*\\
Scenario 2-3 & 8 &  1500  &  0.0137   & 5.3863&  0.8\\
Scenario 3-1 & 10 & 2000 & 0.0074 &  2.3604 & 0.2\\
Scenario 3-2 & 10 & 2000 & 0.0029 &  0.54180 & 0.5*\\
Scenario 3-3 & 10 & 2000 & 0.0262 &  11 & 0.8\\
%Scenario 4-1 & 12  & 2500 & 0.8002 &  0.3073&   0.2079\\
%Scenario 4-2 & 12  & 2500 & 0.8002 &  0.3073&   0.2079\\
%Scenario 4-3 & 12  & 2500 & 0.8002 &  0.3073&   0.2079\\
 \hline
\end{tabular}
\vspace{0.4cm}
\caption* {{Table 1: Comparison between different scenarios of various dimensions and sample sizes in terms of parameter $\alpha$. We applied the FR approach to estimate the GMI ($I_\alpha$) with $\alpha=0.2,0.5,0.8$. We varied dimension $d=6,8,10$ and sample size $n=1000,1500,2000$ in each scenario. We observe that for $\alpha=\{0.2,0.5,0.8\}$, the GMI estimator provides less MSE when $\alpha=0.5$. } }
\end{center}
\vspace{-0.7cm}
\end{table}
\begin{figure}[h]
\centering
    \includegraphics[width=0.6\textwidth]{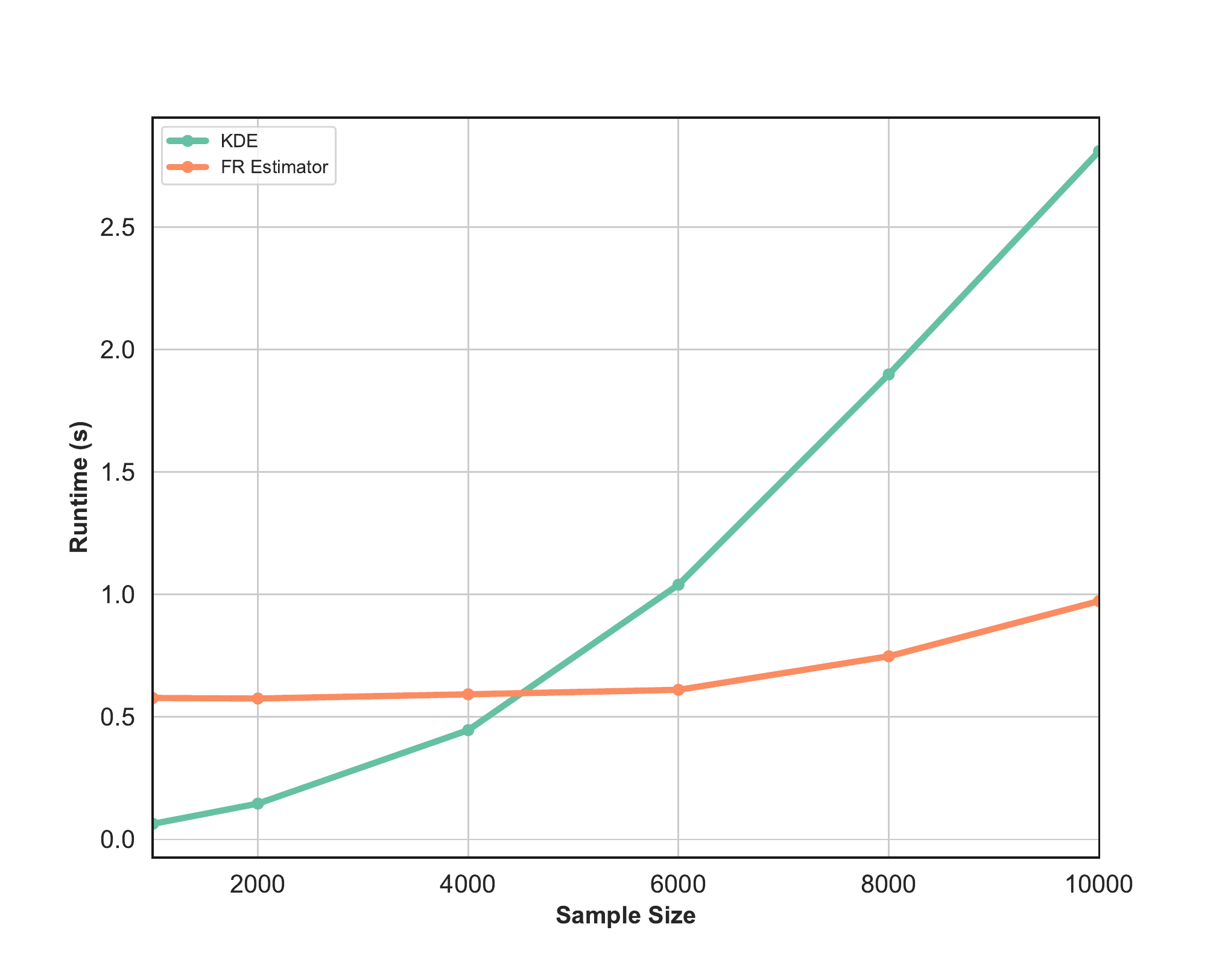}
  \label{timevsN}
  \caption{Runtime of KDE approach and proposed FR estimator of GMI measure vs sample size. The proposed GMI estimator achieves significant speedup, while for small sample size, the KDE method becomes overly fast. Note that in this experiment we have generated our sample from standard Gaussian distribution.}
\end{figure}

Finally Fig. 5 shows the runtime (KDE - FR methods) as a function of sample size $n$. We vary sample size in the range $[10^3,10^4]$. Observe that for smaller number of samples the KDE method is slightly faster but as $n$ becomes large we see significant relative speedup of the proposed FR method.

\section{Conclusion}
In this paper we have proposed a new measure of mutual information, called Geometric MI (GMI), that is related to the Henze-Penrose divergence. The GMI can be viewed as dependency measure that is the limit of the Friedman-Rafsky test statistic that depends on the minimal spanning tree over all data points. We established some properties of the GMI in terms of convexity/concavity, chain rule, and analogous to a data processing inequality. 
A direct estimator of the GMI was introduced that uses random permutations of observed relationships between variables in the multivariate samples.  An explicit form for the MSE convergence rate bound was derived that depends on a free parameter called the proportionality parameter. An asymptotically optimal form for this free parameter was given that minimizes the MSE convergence rate. Simulation studies were performed that illustrate and verify the theory.

\section*{Acknowledgments}

The work presented in this paper was partially supported by ARO grant W911NF-15-1-0479 and DOE grant de-na0002534. The authors would like to thank Brandon Oselio for the useful comments. 
%\appendices
\section{Appendices}

We organize the appendices as the following: Theorem \ref{thm:convex/concave} which establishes convexity/concavity property is proved in Appendix A. Appendix B and C are arranged to establish the inequality (\ref{eq1:thm22}) and (\ref{RHS:data.Proc.}) for given $p\in(0,1)$, respectively. In Appendix D, we first prove that the set $\widetilde{\mathcal{Z}}_{n''}$ which is randomly generated from original dependent data, contains independent samples asymptotically. Later by using the generated independent sample $\widetilde{\mathcal{Z}}_{n''}$ we show that for given $\alpha$ the FR estimator of the GMI given in Algorithm 1 intends to $I_\alpha$ approximately. Appendix E is dedicated to the Theorem \ref{th.Optimized:Bias}. A full discussion on the proportionality parameter ($\alpha$) optimization strategy is provided in Appendix F. 

\subsection{Appendix A: Theorem \ref{thm:convex/concave}}\label{conc.conx}
\begin{proof}
The proof is similar to the result for standard (Shannon) mutual information. However, we require the following lemma, proved in analogue manner as the log-sum inequality:
\begin{Lemma}\label{ApD:lem1}
For non-negative real numbers $\alpha_1,\dots,\alpha_n$ and $\beta_1,\dots,\beta_n$, given $p\in(0,1)$, $q=1-p$, the following holds
\begin{equation*} \diy\sum\limits_{i=1}^n \alpha_i\;\left(p\left(\frac{\beta_i}{\alpha_i}\right)+q\right)^{-1}\geq \diy\sum\limits_{i=1}^n \alpha_i \left(p\left(\frac{\sum\limits_{i=1}^n \beta_i}{\sum\limits_{i=1}^n \alpha_i}\right)+q\right)^{-1}. \end{equation*}
Notice this follows by using the convex function $u(y)=\diy y^2\big/(p+q\;y)$ for any $p\in(0,1)$, $q=1-p$, and the Jensen inequality.
\end{Lemma}
Define the shorthand $\int\limits_{\bx}$, $\int\limits_\by$, and $\int\limits_{\bx\by}$ for $\int \rd\bx$, $\int\rd\by$ and $\iint\;\rd\bx\rd\by$, respectively. To prove part (i) of the Theorem \ref{thm:convex/concave}, we represent the LHS of (\ref{eq1:thm.COn.PMI}) as:
\begin{equation*} \begin{array}{l}
\tilde{I}_p\big(\lambda_1 f_{Y|X}g_X+\lambda_2 f_{Y|X} h_X\big)=\diy 1-\diy\int_{\bx \by}\left(\lambda_1 f_{Y|X}g_X+\lambda_2 f_{Y|X} h_X\right)\times\\
\\
 \diy \left[p\diy\frac{\lambda_1 f_{Y|X}g_X+\lambda_2 f_{Y|X} h_X}{\Big(\int_{\bx} \lambda_1 f_{Y|X}g_X+\lambda_2 f_{Y|X} h_X\Big)\Big(\int_{\by} \lambda_1 f_{Y|X}g_X+\lambda_2 f_{Y|X} h_X\Big)}+q\right]^{-1}\\
\\
=\diy 1-\int_{\bx \by}\left(\lambda_1 f_{Y|X}g_X+\lambda_2 f_{Y|X} h_X\right)\left[p\diy\frac{f_{Y|X}}{\Big(\int_{\bx} \lambda_1 f_{Y|X}g_X+\lambda_2 f_{Y|X} h_X\Big)}+q\right]^{-1}.\end{array}\end{equation*}
Furthermore, the RHS of (\ref{eq1:thm.COn.PMI}) can be rewritten as
\begin{equation*}\begin{array}{l}
\lambda_1 \tilde{I}_p( f_{Y|X}g_X)+\lambda_2 \tilde{I}_p(f_{Y|X}h_X)\\
\\
=\diy 1-\diy\int\limits_{\bx\by}\left(\lam_1 f_{Y|X} g_X \left[p\frac{f_{Y|X}g_X}
{\big(\int\limits_{\bx}f_{Y|X}g_X\big)\big(\int\limits_{\by}f_{Y|X}g_X\big)}+q\right]^{-1}
+\diy\lam_2 f_{Y|X} h_X \left[p\frac{f_{Y|X}h_X}
{\big(\int\limits_{\bx}f_{Y|X}h_X\big)\big(\int\limits_{\by}f_{Y|X}h_X\big)}+q\right]^{-1}\right)\\
\\
\qquad= 1- \diy\int\limits_{\bx\by}\left(\lam_1f_{Y|X}g_X \left[p\frac{f_{Y|X}}{\int\limits_{\bx}f_{Y|X}g_X}+q\right]^{-1}
+\diy\lam_2 f_{Y|X} h_X\left[p\frac{f_{Y|X}}{\int\limits_{\bx}f_{Y|X}h_X}+q\right]^{-1}\right).
\end{array}\end{equation*}
Thus, in order to prove $\hbox{LHS}\geq \hbox{RHS}$, we use the inequality below:
\begin{equation*}\begin{array}{l} \diy \left(\lam_1f_{Y|X}g_X+\lam_2 f_{Y|X}h_X\right)
\left[p\frac{f_{Y|X}}{\int\limits_{\bx}\lam_1f_{Y|X}g_X+\lam_2 f_{Y|X}h_X}+q\right]^{-1}\\
\\
\quad\qquad \leq \diy \left(\lam_1 f_{Y|X}g_X\right)\left[p\frac{\pi}{\int\limits_{\bx}f_{Y|X}g_X}+q\right]^{-1} +\diy \left(\lam_2 f_{Y|X}h_X\right)\left[p\frac{\pi}{\int\limits_{\bx}f_{Y|X}h_X}+q\right]^{-1}.\end{array}\end{equation*}
In Lemma \ref{ApD:lem1}, let
\begin{equation*}\begin{array}{ccl}\diy\alpha_1=\diy\frac{\lam_1\left(\int\limits_{\bx}f_{Y|X}g_X\right)\left(\lam_1f_{Y|X}g_X+\lam_2f_{Y|X}h_{X}\right)}{\int\limits_{\bx}\lam_1f_{Y|X}g_X+\lam_2f_{Y|X}h_{X}},\\
\\
\diy\alpha_2=\diy\frac{\lam_2\left(\int\limits_{\bx}f_{Y|X}h_X\right)\left(\lam_1f_{Y|X}g_X+\lam_2f_{Y|X}h_{X}\right)}{\int\limits_{\bx}\lam_1f_{Y|X}g_X+\lam_2f_{Y|X}h_{X}},
\end{array}\end{equation*}
and for $i=1,2$, 
\begin{equation*}
\beta_i=\diy \frac{\lambda_i f_{Y|X}\Big(\lam_1f_{Y|X}g_X+\lam_2f_{Y|X}h_{X}\Big)}{\int\limits_{\bx}\lam_1f_{Y|X}g_X+\lam_2f_{Y|X}h_{X}}.\end{equation*}
Then the claimed assertion (i) is obtained. Part (ii) follows by convexity of $D_p$ and the following expression: 
\begin{equation*} \begin{array}{l}
\tilde{I}_p\big(\lambda_1 g_{Y|X}f_X+\lambda_2 h_{Y|X} f_X\big)\\
\\
\qquad=\diy D_p\bigg(\lam_1f_X g_{Y|X}+\lam_2f_X h_{Y|X}\;, \bigg(\int\limits_{\bx}\lam_1f_X g_{Y|X}+\lam_2f_X h_{Y|X}\bigg)\diy \bigg(\int\limits_{\by}\lam_1\phi\pi_1+\lam_2\phi\pi_2\bigg)\bigg)\\
\qquad=\diy D_p\bigg(\lam_1f_X g_{Y|X}+\lam_2f_X h_{Y|X}\; , f_X\bigg(\int\limits_{\bx}\lam_1f_X g_{Y|X}+\lam_2f_X h_{Y|X}\bigg)\\
\qquad= \diy D_p\bigg(\lam_1f_X g_{Y|X}+\lam_2f_X h_{Y|X}\; , \diy \lam_1\Big(\int\limits_{\bx}f_X g_{Y|X}\Big)\Big(\int\limits_{\by}f_X g_{Y|X}\Big)+\diy \lam_2 \Big(\int\limits_{\bx}f_X h_{Y|X}\Big)\Big(\int\limits_{\by}f_X h_{Y|X}\Big)\bigg).\end{array}\end{equation*}
Therefore the claim in (\ref{eq2:thm.2.1}) is proved. 
\end{proof}
%\hfill $\quad$ $\blacksquare$
%-----------------------------------------------------------
\subsection{Appendix B: Theorem \ref{thm:Chain.0}}\label{Proof:thm:Chain.0}
\begin{proof}
We prove part (i) and the second part (ii) is shown by repetition method. To show (\ref{eq1:thm22}), we use the inequality below. Given $p\in(0,1)$ and $q=1-p$, we can easily check that for positive $t>q$, $s>q$, such that $t,s\neq 1$:
$$(t+s)\left[(ts)+q(1-t-s)\right]-p (ts)\geq 0.$$
This implies
$$p\Big(\diy\frac{t-q}{p}\Big)\Big(\frac{s-q}{p}\Big)+q\geq \diy\frac{t\;s}{t+s},$$
By substituting
\beqq \diy\frac{f_{X_1Y}(x_1,y)}{f_{X_1}(x_1)\;f_Y(y)}=\diy\frac{t-q}{p},\;\;\;\; \diy\frac{f_{X_2Y|X_1}(x_2,y|x_1)}{f_{X_2|X_1}(x_2|x_1)\;f_{Y|X_1}(y|x_1)}=\diy\frac{s-q}{p},\eeqq
we get
\begin{equation} \begin{array}{l}\diy \left(p\;\frac{f_{X_1X_2Y}(x_1,x_2,y)}{f_{X_1X_2}(x_1,x_2)\;f_Y(y)}+q\right)^{-1}\leq \left(p\;\frac{f_{X_1Y}(x_1,y)}{f_{X_1}(x_1)\;f_Y(y)}+q\right)^{-1}\\
\\
\qquad\qquad+\diy\left(p\;\frac{f_{X_2Y|X_1}(x_2,y|x_1)}{f_{X_2|X_1}(x_2|x_1)\;f_{Y|X_1}(y|x_1)}+q\right)^{-1}.\end{array}\end{equation}
Consequently
\beq \begin{array}{l} I_p(X_1,X_2;Y)\geq I_p(X_1;Y)
 -\diy \bbE_{f}\left[\left(p\;\frac{f_{X_2Y|X_1}(x_2,y|x_1)}{f_{X_2|X_1}(x_2|x_1)\;f_{Y|X_1}(y|x_1)}+q\right)^{-1}\right].\end{array}\eeq
Here $f$ is the joint PDF of random vector $(X_1,X_2,Y)$. This completes the proof of part (i). 
\end{proof}
%-------------------------------------------
%\vspace{-1cm}
\subsection{Appendix C: Proposition \ref{prop.1}}\label{Proof:prop.1}
\begin{proof}
Recall the Theorem \ref{thm:Chain.0}, part (i). First from $\BX\rightarrow \BY\rightarrow \BZ$ we have $f_{XYZ}=f_{XY} f_{Z|Y}$ and then by applying the Jensen inequality, we can write

\begin{equation}\label{Eq:thm.data.Proc.}\begin{array}{l}
I_p(\BX;\BY)=I_p(\BX;\BY,\BZ)\;\;\;\;\;\;\;\hbox{and}\\
\\
I_p(\BX;\BY,\BZ)\geq I_p(\BZ;\BX)-\bbE\bigg[\Big(p\;\diy\pi(\BX,\BY,\BZ)+q\Big)^{-1}\bigg]\\
\\
\qquad\qquad\qquad\geq I_p(\BZ;\BX)-\bigg(p\;\bbE\Big[\diy\pi(\BX,\BY,\BZ)\Big]+q\bigg)^{-1}, \end{array}\end{equation}
where 
$$\pi(\bx,\by,\bz)=\diy\frac{f_{YX|Z}(\by,\bx|\bz)}{f_{Y|Z}(\by|\bz) f_{X|Z}(\bx|\bz)}.$$ 
Now by Markovian property we can immediately simplify the last line of  (\ref{Eq:thm.data.Proc.}) to the RHS in (\ref{RHS:data.Proc.}).

Furthermore we can easily show that if $\BX\rightarrow \BZ\rightarrow \BY$, we have $f_{XYZ}=f_{ZX}f_{Y|Z}$ and therefore $I_p(\BZ;\BX)=I_p(\BX;\BY,\BZ)$. This together with (\ref{Eq:thm.data.Proc.}) proves that under both conditions $\BX\rightarrow \BY\rightarrow \BZ$ and $\BX\rightarrow \BZ\rightarrow \BY$, the equality $I_p(\BX;\BY)=I_p(\BZ;\BX)$ holds true. 
%Similarly if $\BY\rightarrow \BZ\rightarrow \BX$, we have $f_{XYZ}=f_{ZX}f_{Y|Z}$ and therefore $I_p(\BZ;\BX)=I_p(\BX;\BY,\BZ)$. This together with (\ref{Eq:thm.data.Proc.}) proves that under both conditions $\BX\rightarrow \BY\rightarrow \BZ$ and $\BY\rightarrow \BZ\rightarrow \BX$, the equality $I_p(\BX;\BY)=I_p(\BZ;\BX)$ again holds true. 
\end{proof}

%-----------------------------------------------------
\subsection{Appendix D: Theorem \ref{FR:approximation}}\label{Proof.FR:approximation}
\begin{proof}
\def\bbA{\mathbb{A}} \def\bbB{\mathbb{B}}
In this appendix, we first derive two required Lemmas \ref{lem:Indpendent.points} and \ref{lem2:Indpendent.points} below:
\begin{Lemma} \label{lem:Indpendent.points}
Consider random vector $\BZ=(\BX,\BY)$ with joint probability density function (pdf) $f_{XY}$. Let $\mathfrak{Z}_n=\{\bz_1,\ldots,\bz_n\}=\{(\bx_i,\by_i)\}_{i=1}^n$ be a set of samples with pdf $f_{XY}$. Let ${\mathfrak{Z}}'_{n'}$ and ${\mathfrak{Z}}''_{n''}$ be two distinct subsets of $\mathfrak{Z}_n$ such that $n'+n''=n$ and sample proportion is $\alpha=n'/n$ and $\beta=1-\alpha$. Next, let $\widetilde{\mathfrak{Z}}_{n''}=\{\widetilde{\bz}_1,\ldots,\widetilde{\bz}_{n''}\}$ be a set of pairs such that $\widetilde{\bz}_{k}=(\bx_{i_k},\by_{j_k})$, $k=1,\dots,n''$ are selected at random from ${\mathfrak{Z}}'_{n''}$. Denote $\widetilde{\BZ}=(\widetilde{\BX},\widetilde{\BY})$ as the random vector corresponding to samples in $\widetilde{\mathfrak{Z}}_{n''}$. Then as $n\rightarrow\infty$ such that $n''$ also grows in a linked manner that $\beta\neq 0$ then the distribution of $\widetilde{\BZ}$ convergences to $f_X\times f_Y$ i.e. random vectors $\widetilde{\BX}$ and $\widetilde{\BY}$ become independent. 
%$\widetilde{\BZ}_{n''}$ convergences to $\widetilde{\mathbf{Z}}$ in distribution i.e. $f_{X}\times f_{Y}$. This means that for  $\widetilde{\BZ}=(\widetilde{\BX},\widetilde{\BY})$, random vectors $\widetilde{\BX}$ and $\widetilde{\BY}$ are independent. 
\end{Lemma}
\begin{proof} Consider two subsets $\bbA,\;\bbB\subset \bbR^n$, then we have
\begin{equation*}\begin{array}{l}\diy P(\widetilde{\BX}\in \bbA, \widetilde{\BY}\in \bbB)=\bbE\left[\mathbf{I}_{\bbA}({\widetilde{\BX}}).\;\mathbf{I}_{\bbB}({\widetilde{\BY}})\right]
=\diy \bbE\left[\diy\sum\limits_{i,j} \mathbf{I}_{\bbA}(\BX_i).\; \mathbf{I}_{\bbB}(\BY_j). \;P\big((\widetilde{\BX},\widetilde{\BY})=(\BX_i,\BY_j)\big|\BZ_n\big)\right].\end{array}\end{equation*}
Here $\mathbf{I}_{\bbA}$ stands for the indicator function. Note that 
\begin{equation*} P\Big((\widetilde{\BX},\widetilde{\BY})=(\BX_i,\BY_j)\big|\BZ_n\Big)=\diy\frac{1}{{n''}^2},\end{equation*}
and $\BX_i$ and $\BY_j$, $i\neq j$ are independent, therefore
\begin{equation*}\begin{array}{l}  P(\widetilde{\BX}\in \bbA, \widetilde{\BY}\in \bbB)
=\diy\frac{1}{{n''}^2}\diy\sum\limits_{i\neq j} P(\BX_i\in\bbA)P(\BY_j\in \bbB)
+\diy\frac{1}{{n''}^2}\sum\limits_{i=1}^n P(\BX_i\in\bbA,\BY_i\in\bbB)\\
%&=& \diy\frac{n(n-1)}{n^2}P(\BX_i\in\bbA)P(\BY_j\in \bbB)+\diy\frac{1}{n}\sum\limits_{i=1}^{n''} P(\BX_i\in\bbA,\BY_i\in\bbB)\\
\quad= \diy P(\BX_i\in\bbA)P(\BY_j\in \bbB)
+\diy\frac{1}{n''}\bigg\{P(\BX_i\in\bbA,\BY_i\in\bbB)-P(\BX_i\in\bbA)P(\BY_i\in\bbB)\bigg\},\end{array}\end{equation*}
this implies that
\begin{equation} \begin{array}{l} \diy \Big| P(\widetilde{\BX}\in \bbA, \widetilde{\BY}\in \bbB)-P(\widetilde{\BX}\in\bbA)P(\widetilde{\BY}\in\bbB)\Big|
\leq \diy\frac{1}{n''} \iint \Big|f_{XY}(\bx,\by)-f_{X}(\bx) f_{Y}(\by)\Big|\;\rd\bx\;\rd\by. \end{array}\end{equation}
On the other hand, we know that $n''=\beta\;n$, so we get
\begin{equation}\label{eq1:lema1} \begin{array}{l}\diy \Big| P(\widetilde{\BX}\in \bbA, \widetilde{\BY}\in \bbB)-P(\widetilde{\BX}\in\bbA)P(\widetilde{\BY}\in\bbB)\Big|
\leq \diy\frac{1}{\beta\;n} \iint \Big|f_{XY}(\bx,\by)-f_{X}(\bx) f_{Y}(\by)\Big|\;\rd\bx\;\rd\by.\end{array} \end{equation}
From (\ref{eq1:lema1}), we observe that when $\beta$ takes larger values the bound becomes tighter. So if $n\rightarrow \infty$ such that $n''$ also becomes large enough in a linked manner that $\beta\neq 0$ then the RHS in (\ref{eq1:lema1}) tends to zero. This implies that $\widetilde{\BX}$ and $\widetilde{\BY}$ become independent when $n\rightarrow \infty$.
\end{proof}
An immediate result of Lemma \ref{lem:Indpendent.points} is the following:
\begin{Lemma} \label{lem2:Indpendent.points}
For given random vector $\BZ_n=(\BX_n,\BY_n)$ from joint density function $f_{XY}$ and with marginal density functions $f_{X}$ and $f_{Y}$
set $\widetilde{\mathfrak{Z}}_{n''}=\big\{\widetilde{\bz}_1,\dots, \widetilde{\bz}_{n''}\big\}$ be realization of random vector $\widetilde{\BZ}$ as in Lemma \ref{lem:Indpendent.points} with parameter $\beta=n''/n$.  Then for given points of $\widetilde{\mathfrak{Z}}_{n''}$ at $\widetilde{\bz}=(\widetilde{\bx},\widetilde{\by})$, we have
\begin{equation} \Big| f_{\widetilde{Z}}(\widetilde{\bx},\widetilde{\by})-f_{X}(\widetilde{\bx}) f_{Y}(\widetilde{\by})\Big|=O\left(\diy\frac{1}{\beta n}\right).\end{equation}
\end{Lemma}
Now, let us get back to our main goal which is the proof of assertion (\ref{app:MIP}). Consider two subsets $\BZ'_{n'}$ and $\widetilde{\BZ}_{n''}$ as described in Subsection \ref{subsec:2.1} . Assume that the components of sample $\widetilde{\BZ}_{n''}$ follow density function $\widetilde{f}_{\widetilde{X}\widetilde{Y}}$. Therefore by owing to Lemma \ref{lem:Indpendent.points} and \ref{lem2:Indpendent.points}, when $n\rightarrow\infty$ then $\widetilde{f}_{\widetilde{X}\widetilde{Y}}\rightarrow f_{X} f_{Y}$.  Let $M_{n'}$ and $N_{n''}$ be Poisson variables with mean $n'$ and $n''$ independent of one another and $\{{\BZ}'_i\}$ and $\{\widetilde{\BZ}_j\}$.  Assume two Poisson processes $\mathfrak{Z}'_{n'}=\big\{{\BZ}'_1,\dots, {\BZ}'_{M_{n'}}\big\}$ and $\widetilde{\mathfrak{Z}}_{n''}=\big\{\widetilde{\BZ}_1,\dots,\widetilde{\BZ}_{N_{n''}}\big\}$, with the FR statistic $\mathfrak{R}'_{n',n''}$. So by owing to the arguments in \cite{HP}, \cite{YNMH} we shall prove the following:
\begin{equation*} \diy\frac{\bbE\left[\mathfrak{R}'_{n',n''}\right]}{n'+n''}\rightarrow 2\alpha\beta \;\diy\iint\diy\frac{f_{X,Y}(\bx,\by) f_{X}(\bx) f_{Y}(\by)}{\alpha f_{XY}(\bx,\by)+\beta f_{X}(\bx) f_{Y}(\by)}\;\rd\bx\;\rd\by. \end{equation*}
This is because of $\big|\mathfrak{R}'_{n',n''}-\mathfrak{R}_{n',n''}\big|\leq K_d\Big(|M_{n'}-n'|+|N_{n''}-n''|\Big)$, where $K_d$ is a constant defined in Lemma 1, \cite{HP} and $n'+n''=n$, then $(n'+n'')^{-1}\;\bbE\big|\mathfrak{R}'_{n',n''}-\mathfrak{R}_{n',n''}\big|\rightarrow 0$ as $n\rightarrow \infty$.  
 Let $\BW_1^{n',n''}, \BW_2^{n',n''},\dots$ be independent variables with common density 
 $$\phi_{n',n''}(\bx,\by)=\Big(n' f_{XY}(\bx,\by)+n''\widetilde{f}_{\widetilde{X}\widetilde{Y}}(\bx,\by)\Big)\big/(n'+n''),$$
 for $(\bx,\by)\in \bbR^d\times\bbR^d$.  Let $L_{n',n''}$ be an independent Poisson variable with mean $n'+n''$.  Let $\mathfrak{F}'_{n',n''}=\left\{\BW_1^{n',n''},\dots,\BW_{L_{n',n''}}^{n',n''}\right\}$ a nonhomogeneous Poisson process of rate $ n'\; f_{XY}+n''\; \widetilde{f}_{\widetilde{X}\widetilde{Y}}$. Assign mark $1$ to a point in $\mathfrak{F}'_{n',n''}$ with probability
$$n'\; f_{XY}(\bx,\by)\Big/\left(n'\; f_{XY}(\bx,\by)+n''\; \widetilde{f}_{\widetilde{X}\widetilde{Y}}(\bx,\by)\right),$$
and mark 2 otherwise, having the FR test statistic $\widetilde{\mathfrak{R}}_{n',n''}$  which by the marking theorem \cite{Ki}, \cite{HP}, has a same distribution as $\mathfrak{R}'_{n',n''}$.  Given points of $\mathfrak{F}'_{n',n''}$ at $\bz'=(\bx',\by')$ and $\bz''=(\bx'',\by'')$, the probability that they have different marks is given by (\ref{setcounter23}). 
%\begin{figure*}[h]
%\hrulefill
\vspace*{4pt}
\normalsize
\begin{equation}\label{setcounter23}
g_{n',n''}(\bz',\bz'')=\diy\frac{n'\;f_{XY}(\bx',\by')\; n''\; \widetilde{f}_{\widetilde{X},\widetilde{Y}}(\bx'',\by'')+ n''\;\widetilde{f}_{\widetilde{X},\widetilde{Y}}(\bx',\by')\; n'\; f_{XY}(\bx'',\by'')}{\left(n'\;f_{XY}(\bx',\by')+n''\; \widetilde{f}_{\widetilde{X},\widetilde{Y}}(\bx',\by')\right)\left(n'\; f_{XY}(\bx'',\by'')+n''\; \widetilde{f}_{\widetilde{X},\widetilde{Y}}(\bx'',\by'')\right)},
\end{equation}
Set 
\begin{equation}\label{setcounter23.2} g(\bz',\bz'')=\diy\frac{\alpha \beta\left(f_{XY}(\bx',\by')f_{X}(\bx'') f_{Y}(\by'')+f_{X}(\bx')f_{Y}(\by') f_{XY}(\bx'',\by'')\right)}{\bigg(\alpha\; f_{XY}(\bx'',\by'')+\beta f_X(\bx'')f_Y(\by'')\bigg)\bigg(\alpha f_{XY}(\bx',\by')+\beta f_{X}(\bx')f_{Y}(\by')\bigg)},\end{equation}
%\setcounter{equation}{\value{MYtempeqncnt}}
%\hrulefill
%\end{figure*}
%-------------------------------------------------------------
then
\begin{equation}\begin{array}{l} \label{eq:A.1} \bbE\Big[\widetilde{\mathfrak{R}}'_{n',n''}|\mathfrak{F}'_{n',n''}\Big]
=\diy \mathop{\sum\sum}_{  i<j\leq L_{n',n''}} g_n(\BW_i^{n',n''},\BW_j^{n',n''})\mathbf{I}_{\mathfrak{F}'_{n',n''}}(\BW_i^{n',n''},\BW_j^{n',n''}).\end{array} \end{equation}
Now recall (\ref{setcounter23.2}). We observe that $g_{n',n''}(\bz',\bz'')\rightarrow  g(\bz',\bz'')$. Going back to (\ref{eq:A.1}), we can write
\begin{equation}\begin{array}{l} \bbE\Big[\widetilde{\mathfrak{R}}'_{n',n''}\Big]
= \diy \mathop{\sum\sum}_{  i<j\leq L_{n',n''}} g_{n',n''}(\BW_i^{n',n''},\BW_j^{n',n''})\mathbf{I}_{\mathfrak{F}'_{n',n''}}(\BW_i^{n',n''},\BW_j^{n',n''})
+o(n'+n'').\end{array}\end{equation}
Consider the non-Poisson process 
$$\mathfrak{F}_{n',n''}=\left\{\BW_1^{n',n''},\dots,\BW_{n',n''}^{n'+n''}\right\}.$$
 So by the fact that $\bbE\Big[M_{n'}+N_{n''}-(n'+n'')\Big]=o(n'+n'')$, we have
\begin{equation}\begin{array}{l}\diy  \bbE\Big[\widetilde{\mathfrak{R}}'_{n',n''}\Big]
=\diy \mathop{\sum\sum}_{  i<j\leq n'+n''} g_{n',n''}(\BW_i^{n',n''},\BW_j^{n',n''})\mathbf{I}_{\mathfrak{F}_{n',n''}}(\BW_i^{n'},\BW_j^{n''})
+o(n'+n'').\end{array}\end{equation}
Introduce 
$$\phi(\bx,\by)=\alpha f_{XY}(\bx,\by)+\beta f_{X}(\bx) f_{Y}(\by).$$
 Then $\phi_{n',n''}(\bx,\by)\rightarrow \phi(\bx,\by)$ uniformly because of $\diy{n'}/{n}\rightarrow \alpha$ and $\diy{n''}/{n}\rightarrow \beta$. Thus by using Proposition 1 in \cite{HP}, we get
\begin{equation}\begin{array}{l} \diy\frac{\bbE\left[\widetilde{\mathfrak{R}}'_{n',n''}\right]}{n}\rightarrow \int g(\bz,\bz)\phi(\bz)\rd\bz
=\diy\iint\frac{2\alpha\beta f_{XY}(\bx,\by)f_{X}(\bx)f_{Y}(\by)}{\alpha f_{XY}(\bx,\by)+\beta f_{X}(\bx) f_{Y}(\by)}\;\rd\bx\;\rd\by. \end{array}\end{equation}
So,  we conclude the proof.
\end{proof}
%--------------------------------------------------------
\subsection{Appendix E: Theorem \ref{th.Optimized:Bias}}\label{Proof:th.Optimized:Bias}
\begin{proof}
We begin by providing a family of bias rate bound for the FR estimator $\mathfrak{R}_{n',n''}$ in terms of a parameter $l$. Then by plugging the optimal $l$, we prove the bias rate bound given in (\ref{Optimized:Bias}).
\begin{Theorem}\label{main:Bias}
Let $\mathfrak{R}_{n',n''}:=\mathfrak{R}(\mathfrak{Z}_{n'},\mathfrak{Z}_{n''})$ be the FR test statistic. Then a bound on the bias rate of the $\mathfrak{R}_{n',n''}$ estimator for $0<\eta\leq 1$, $d\geq 2$ is given by
\begin{equation}\label{bound:Bias} \begin{array}{l}\diy \bigg|\frac{\bbE\left[\mathfrak{R}_{n',n''}\right]}{n}-2\alpha \beta\iint \frac{f_{XY}(\bx,\by) f_{X}(\bx) f_{Y}(\by)}{\alpha f_{XY}(\bx,\by)+\beta f_{X}(\bx)f_{Y}(\by)}\;\rd\bx \bigg|\\
\\
\leq 
O\left(l^d(n)^{-\eta/d}\right) +O\left(l^{-d\eta}\right)+O\left(l^{d}\beta^{-1}n^{-1}\right)+O\left(c_d 2^d n^{-1}\right),
\end{array}\end{equation}
where $0<\eta\leq 1$ is the H\"{o}lder smoothness parameter and $c_d$ is a constant depending only on $d$.\\
Set
\begin{equation*} \begin{array}{ccl}\alpha_i=\alpha n a_i l^{d}\;\left(1-a_i l^{-d}\right)+{\left(\alpha n\right)}^2 a_i^2, \\
\\
 \beta_i=\beta n b_i l^d\;\left(1-b_i l^{-d}\right)+{\left(\beta n\right)}^2 b_i^2 .\end{array}\end{equation*}
 and
\begin{equation} \label{eq:Def:A} \diy \mathcal{A}_{f,n}^{\beta,\alpha}(\bx,\by)=\diy\frac{2f_{XY}(\bx,\by)\left(f_{X}(\bx)f_{Y}(\by)+\delta_f\big/(\beta n)\right)\left(f_{XY}(\bx,\by)\sqrt{\alpha }+\big(f_{X}(\bx)f_{Y}(\by)+\delta_f\big/(\beta n))\sqrt{\beta}\right)}{a_i^2 l^{-d}\left(\alpha f_{XY}(\bx,\by)+\beta  \Big(f_{X}(\bx)f_{Y}(\by)+\delta_f\big/(\beta n)\Big)\right)^2},\end{equation}
where 
\begin{equation}\label{delta.def}
\delta_f=\diy\iint\Big|f_{XY}(\bx,\by)-f_{X}(\bx)f_{Y}(\by)\Big|\;\rd\bx \rd\by,\end{equation}
A more explicit form for the bound on the RHS is given below:
\begin{equation} \label{warpi.func}
\begin{array}{cl}\Delta(\alpha,f_{XY},f_{X}f_{Y}):= \diy c_2 l^d (n)^{-1}+c_d 2^d(n)^{-1} +O\Big(l^d(n)^{-\eta/d}\Big)+O\Big(l^d(n)^{-1/2}\Big)\\
\\
+O\left(c_d(n)^{-1/2}\right)+2c_1l^{d-1}(n)^{(1/d)-1}+\diy \delta_f ((\beta n)^{-1}) \iint \diy\frac{2\alpha \beta f_{XY}(\bx,\by)}{\alpha f_{XY}(\bx,\by)+\beta f_{X}(\bx)f_{Y}(\by)}\;\rd\bx\rd\by \\
\\
+\diy (n)^{-1} \sum\limits_{i=1}^M2\iint \diy f_{XY}(\bx,\by) \Big( f_{X}(\bx)f_{Y}(\by)+\delta_f\big/(\beta n)\Big)\diy \Big(\alpha_i \beta_i\Big(\alpha n a_i l^{-d} f^2_{XY}(\bx,\by)\\
\\
\quad+\beta n b_i l^{-d}  \big( f_{X}(\bx)f_{Y}(\by)+\delta_f\big/(\beta n)\big)^2\Big) \Big)^{1/2}
\Big/\big(\alpha n a_i f_{XY}(\bx,\by)+\beta n  b_i  f_{X}(\bx)f_{Y}(\by)\big)^2 \;\rd \bx\rd\by\\
\\
\quad +\diy (n)^{-1}  \sum\limits_{i=1}^M O(l) \iint l^d (a_i)^{-1}\diy\frac{2 f_{XY}(\bx,\by)\Big(f_{X}(\bx)f_{Y}(\by)+\delta_f\big/(\beta n)\Big)}{\alpha   f_{XY}(\bx,\by)+\beta  f_{X}(\bx)f_{Y}(\by)}\;\rd\bx\rd\by\\
\\
\quad+\diy (n)^{-3/2} \sum\limits_{i=1}^M O(l)\iint l^{-d/2}\sqrt{b_i}\; \mathcal{A}_{f,n}^{\beta,\alpha}(\bx,\by)\;\rd\bx\rd\by.
\end{array}
\end{equation}
\end{Theorem}
\begin{proof}
Consider two Poisson variables $M_{n'}$ and $N_{n''}$ with mean $n'$ and $n''$ respectively and independent of one another and $\{\BZ'_i\}$ and $\{\widetilde{\BZ}_j\}$. Let $\mathfrak{Z}'_{n'}$ and $\widetilde{\mathfrak{Z}}_{n''}$ be the Poisson processes $\{\BZ'_1,\dots,\BZ'_{M_{n'}}\}$ and $\{\widetilde{\BZ}_1,\dots,\widetilde{\BZ}_{N_{n''}}\}$. Likewise Appendix D, set $\mathfrak{R}'_{n',n''}=\mathfrak{R}(\mathfrak{Z}'_{n'},\widetilde{\mathfrak{Z}}_{n''})$. Applying Lemma 1, and (12) in \cite{HP}, we can write
\beq \label{eq:5.19}\Big|\mathfrak{R}'_{n',n''}-\mathfrak{R}_{n',n''}\Big|\leq c_d \big(|M_{n'}-n'|+|N_{n''}-n''|). \eeq
Here $c_d$ denotes the largest possible degree of any vertex of the MST in $\bbR^d$. Following the arguments in \cite{YNMH}, we have $\bbE\big[\big|M_{n'}-n'\big|\big]=O\big(\;{n'}^{1/2}\big)$ and  $\bbE\big[\big|N_{n''}-n''\big|\big]=O\big(\;{n''}^{1/2}\big)$. Hence
\begin{equation}\label{eqB1:2.23} \diy\frac{ \bbE\Big[\mathfrak{R}_{n',n''}\Big]}{n'+n''}=\diy\frac{\bbE\left[\mathfrak{R}'_{n',n''}\right]}{n'+n''}+O\left(c_d(n'+n'')^{-1/2}\right).
\end{equation}
Next let $n'_i$ and $n''_i$ be independent binomial random variables with marginal densities $B(n',a_il^{-d})$ and $B(n'',b_il^{-d})$ such that $a_i,b_i$ are non-negative constants $a_i\leq b_i$ and $\diy\sum\limits_{i=1}^{l^d}a_i l^{-d}=\diy\sum\limits_{i=1}^{l^d} b_i l^{-d}=1$. Therefore using subadditivity property in Lemma 2.2, \cite{YNMH}, we can write
\begin{equation}\label{Eq1:2:2.19}\begin{array}{l} \diy \bbE\Big[\mathfrak{R}'_{n',n''}\Big]
\leq \diy\sum\limits_{i=1}^M\bbE\bigg[\bbE\Big[\mathfrak{R}'_{n'_i,n''_i}|n'_i,n''_i\Big]\bigg]+2\;c_1\;l^{d-1}(n'+n'')^{1/d},\end{array}\end{equation}
where $ M=l^d$, and $\eta>0$ stands with H\"{o}lder smoothness parameter. Further, for given $n'_i$, $n''_i$,   let $\BW_1^{n'_i,n''_i}, \BW_2^{n'_i,n''_i}, \dots$ be independent variables with common densities for $ (\bx,\by)\in\bbR^d \times\bbR^d$: 
$$g_{n'_i,n''_i}(\bx,\by)=\left(n'_i f_{XY}(\bx,\by)+n''_i \widetilde{f}_{\widetilde{X}\widetilde{Y}}(\bx,\by)\right)\big/(n'_i+n''_i).$$
Denote $L_{n'_i,n''_i}$ be an independent Poisson variable with mean $n'_i+n''_i$  and  $\mathfrak{F}'_{n'_i,n''_i}=\big\{\BW_1^{n'_i,n''_i},\dots, \BW_{L_{n'_i.n''_i}}^{n'_i,n''_i}\big\}$ a non-homogeneous Poisson of rate $n'_i f_{XY}+n''_i \widetilde{f}_{\widetilde{X}\widetilde{Y}}$.  Let $\mathfrak{F}_{n'_i,n''_i}$ be the non-Posisson point process $\big\{\BW_1^{n'_i,n''_i},\dots \BW_{n'_i+n''_i}^{n'_i,n''_i}\big\}$. Assign a mark from the set $\{1,2\}$ to each points of $\mathfrak{F}'_{n'_i,n''_i}$.  Let $\widetilde{\mathfrak{Z}}'_{n'_i}$ be the sets of points marked 1 with each probability $n'_i f_{XY}(\bx,\by)\big/ \big(n'_i f_{XY}(\bx,\by)+n''_i \widetilde{f}_{\widetilde{X}\widetilde{Y}}(\bx,\by)\big)$ and let $\widetilde{\mathfrak{Z}}''_{n''_i}$ be the set points with mark 2. Note that owing to the marking theorem  \cite{Ki}, $\widetilde{\mathfrak{Z}}'_{n'_i}$ and $\widetilde{\mathfrak{Z}}''_{n''_i}$ are independent Poisson processes with the same distribution as $\mathfrak{Z}'_{n'_i}$ and $\widetilde{\mathfrak{Z}}_{n''_i}$ , respectively.  Considering $\widetilde{\mathfrak{R}}'_{n'_i,n''_i}$ as FR test statistic on nodes in $\widetilde{\mathfrak{Z}}'_{n'_i}\cup \widetilde{\mathfrak{Z}}''_{n''_i}$, we have 
\beqq \bbE\Big[\mathfrak{R}'_{n'_i,n''_i}|n'_i,n''_i\Big]=\bbE\Big[\widetilde{\mathfrak{R}}'_{n'_i,n''_i}|n'_i,n''_i\Big].\eeqq
By the fact that $\bbE\big[|M_{n'}+N_{n''}-n'-n''|\big]=O((n'+n'')^{1/2})$, we have
\beqq\begin{array}{l} \diy\bbE\Big[\widetilde{\mathfrak{R}}'_{n'_i,n''_i}|n'_i,n''_i\Big]=\diy\bbE\Big[\bbE\big[\widetilde{\mathfrak{R}}'_{n'_i,n''_i}|\mathfrak{F}'_{n'_i,n''_i}\big]\Big]\\
\\
\qquad=\diy\bbE\left[\diy\mathop{\sum\sum}_{s<j<n'_i+n''_i}P_{n'_i,n''_i}(\BW_s^{n'_i,n''_i},\BW_j^{n'_i,n''_i}) \; \diy \mathbf{1}
\left\{(\BW_s^{n'_i,n''_i},\BW_j^{n'_i,n''_i})\in\mathfrak{F}_{n'_i,n''_i}\right\}\right]+O\left((n'_i+n''_i)^{1/2})\right). 
\end{array}\eeqq
Here $\bz'=(\bx',\by')$, $\bz''=(\bx'',\by'')$, and $P_{n'_i,n''_i}(\bz',\bz'')$ is given in below: 

\beqq\label{setcounter:1}\begin{array}{l} P_{n'_i,n''_i}(\bz',\bz''):=\diy P_{r}\left\{\hbox{mark}\; \bz'\neq \hbox{mark}\;\bz'', (\bz',\bz'')\in \mathfrak{F}_{n'_i,n''_i}\right\}\\
\\
=\diy\frac{n'_if_{XY}(\bx',\by') n''_i\widetilde{f}_{\widetilde{X}\widetilde{Y}}(\bx'',\by'')+n'_i\widetilde{f}_{\widetilde{X}\widetilde{Y}}(\bx',\by')n''_if_{X,Y}(\bx'',\by'')}{\left(n''_i f_{XY}(\bx',\by')+n''_i \widetilde{f}_{\widetilde{X}\widetilde{Y}}(\bx',\by')\right)\left(n'_1f_{XY}(\bx'',\by'')+n''_i \widetilde{f}_{\widetilde{X}\widetilde{Y}}(\bx'',\by'')\right)}.\end{array}\eeqq
Next set
 \beqq\begin{array}{ccl} \alpha_i=n' a_i l^{d}\;(1-a_i l^{-d})+{n'}^2 a_i^2,\;\;\;
 \beta_i=n'' b_i l^d\;(1-b_i l^{-d})+{n''}^2 b_i^2 . \end{array}\eeqq
By owing to the Lemma B.6 in  \cite{YNMH} and applying the analogous arguments, we can write the expression in (\ref{eq0:Apendix.B}):
 \begin{equation}\label{eq0:Apendix.B} \begin{array}{l}
\bbE\Big[{\mathfrak{R}'}_{n',n''}\Big]\leq \diy \sum\limits_{i=1}^M \;a_i b_i l^{-d}\iint\diy\frac{2\;n' n'' f_{XY}(\bx,\by) f_{\widetilde{X}\widetilde{Y}}(\bx,\by)}{n' a_i f_{X,Y}(\bx,\by)+n'' b_i  f_{\widetilde{X}\widetilde{Y}}(\bx,\by)}\;\rd\bx\;\rd\by+2c_1\;l^{d-1}(n'+n'')^{1/d}\\
\\
\quad+ \diy\sum\limits_{i=1}^M2\iint \diy \frac{f_{XY}(\bx,\by)  f_{\widetilde{X}\widetilde{Y}}(\bx,\by)\left(\alpha_i \beta_i\big(n' a_i l^{-d} f^2_{XY}(\bx,\by)+n'' b_i l^{-d}  f^2_{\widetilde{X}\widetilde{Y}}(\bx,\by)\big) \right)^{1/2}}{\left(n' a_i f_{XY}(\bx,\by)+n'' b_i  f_{\widetilde{X}\widetilde{Y}}(\bx,\by)\right)^2}\;\rd \bx\rd\by\\
\\
\quad+\diy\sum\limits_{i=1}^M\bbE_{n'_i,n''_i}\left[(n'_i+n''_i)\; \varsigma_\eta(l,n'_i,n''_i) \right] + O\left(l^d(n'+n'')^{1-\eta/d}\right)
+\diy O\left(l^d(n'+n'')^{1/2}\right),
\end{array}\end{equation}
where
\beqq \varsigma_\eta(l,n'_i,n''_i)=\bigg(O\Big(\frac{l}{n'_i+n''_i}\Big)-\diy\frac{2\;l^{d}}{n'_i+n''_i}\bigg)\diy\int g_{n'_i,n''_i}(\bz')P_{n'_i,n''_i}(\bz',\bz')\;\rd\bz'+O(l^{-d\eta}).\eeqq
Going back to Lemma \ref{lem2:Indpendent.points}, we know that
\beqq f_{\widetilde{X}\widetilde{Y}}(\bx,\by)=f_{X}(\bx) f_{Y}(\by)+O\left(\diy\frac{1}{\beta n}\right).\eeqq
Therefore the first term in RHS of (\ref{eq0:Apendix.B}) turns to be less and equal than
\beqq \begin{array}{l} \diy \sum\limits_{i=1}^M \;a_i b_i l^{-d}\iint\diy\frac{2\;n' n'' f_{XY}(\bx,\by) f_{X}(\bx) f_{Y}(\by)}{n' a_i f_{X,Y}(\bx,\by)+n'' b_i   f_{X}(\bx) f_{Y}(\by)}\;\rd\bx\;\rd\by\\
\\
+\diy \left(\diy\frac{\delta_f}{\beta n}\right) \diy \sum\limits_{i=1}^M \;a_i b_i l^{-d}\iint\diy\frac{2\;n' n'' f_{XY}(\bx,\by)}{n' a_i f_{XY}(\bx,\by)+n'' b_i   f_{X}(\bx) f_{Y}(\by)}\;\rd\bx\;\rd\by,\end{array}\eeqq
and the second term is less and equal than
 \beqq \begin{array}{l}\diy\sum\limits_{i=1}^M2\iint \diy f_{XY}(\bx,\by) \Big( f_{X}(\bx)f_{Y}(\by)+\delta_f\big/(\beta n)\Big)\diy
\bigg(\alpha_i \beta_i\Big(n' a_i l^{-d} f^2_{XY}(\bx,\by)+n'' b_i l^{-d}  \big( f_{X}(\bx)f_{Y}(\by)\\
\qquad \qquad +\diy\delta_f\big/(\beta n)\big)^2\Big) \bigg)^{1/2}
\Big/\left(n' a_i f_{XY}(\bx,\by)+n'' b_i  f_{X}(\bx)f_{Y}(\by)\right)^2\;\rd \bx\rd\by,\end{array}\eeqq
where 
\beqq \delta_f=\iint\Big|f_{XY}(\bx,\by)-f_{X}(\bx)f_{Y}(\by)\Big|\;\rd\bx\rd\by.\eeqq
Recall the definition of the dual MST and FR statistic denoted by $\mathfrak{R}^*_{n',n''}$ from \cite{YNMH}:

 \def \bbF{\mathbb{F}}
 {\bf Definition:}
{\rm (Dual MST, ${\rm MST}^*$ and dual FR statistic $\mathfrak{R}_{m,n}^*$)}  Let $\bbF_i$ be the set of corner points of the subsection $Q_i$ for $ 1\leq i \leq l^d$. Then we define ${\rm MST}^*(\mathfrak{X}_m\cup\mathfrak{Y}_n \cap Q_i)$ as the boundary MST graph of partition $Q_i$ \cite{Yu}, which contains $\mathfrak{X}_m$ and $\mathfrak{Y}_n$ points falling inside the section $Q_i$ and those corner points in $\bbF_i$ which minimize total MST length. Notice it is allowed to connect the MSTs in $Q_i$ and $Q_j$ through points strictly contained in $Q_i$ and $Q_j$ and corner points are taking into account under condition of minimizing total MST length. Another word, the dual MST can connect the points in $Q_i\cup Q_j$ by direct edges to pair to another point in $Q_i\cup Q_j$ or the corner the corner points (we assume that all corner points are connected) in order to minimize the total length. To clarify this, assume that there are two points in $Q_i\cup Q_j$, then the dual MST consists of the two edges connecting these points to the corner if they are closed to a corner point otherwise dual MST consists of an edge connecting one to another.  
%depending on the length of corners from nodes, might  contain the corner points $\bbF_i$ too. 
Further, $\mathfrak{R}_{m,n}^*(\mathfrak{X}_m,\mathfrak{Y}_n \cap Q_i)$ is defined as the number of edges in ${\rm MST}^*$ graph connecting nodes from different samples and number of edges connecting to the corner points. Note that the edges connected to the corner nodes (regardless of the type of points) are always counted in dual FR test statistic $\mathfrak{R}^*_{m,n}$. 

See \cite{YNMH} for more detail about the dual MST and dual FR test statistic. Similar discussion as above and in \cite{YNMH}, consider the Poisson processes samples and  the FR test statistic under the union of samples, denoted by ${\mathfrak{R}'}^*_{n',n''}$, and superadditivity of dual $\mathfrak{R}^*_{n',n''}$,  we have
\begin{equation}\label{eq1:Appendix.B}\begin{array}{l} \bbE\Big[\mathfrak{R}'^*_{n',n''}\Big]
\geq \diy\sum\limits_{i=1}^M a_i\;l^{-d}\iint \diy\frac{2n'n'' f_{XY}(\bx,\by)\left(f_{X}(\bx)f_{Y}(\by)-\delta_f/(\beta n)\right)}{n' f_{XY}(\bx,\by)+n''\left(f_{X}(\bx)f_{Y}(\by)-\delta_f/(\beta n)\right)}\;\rd\bx\rd\by\\
\\
\quad-\diy\sum\limits_{i=1}^M\bbE_{n'_i,n''_i}\Big[(n'_i+n''_i)\; \varsigma_\eta(l, n'_i,n''_i) \Big]-O\left(l^d(n'+n'')^{1-\eta/d}\right)
- O\left(l^d(n'+n'')^{1/2}\right) - c_2\;l^d.\end{array}\end{equation}
The first term of RHS in (\ref{eq1:Appendix.B}) is greater and equal than 
\beqq \begin{array}{l} \diy \iint \diy\frac{2n'n'' f_{XY}(\bx,\by) f_{X}(\bx)f_{Y}(\by)}{n' f_{XY}(\bx,\by)+n'' f_{X}(\bx)f_{Y}(\by)}\;\rd\bx\rd\by
-\diy\frac{\delta_f}{\beta n}  \iint \diy\frac{2n'n'' f_{XY}(\bx,\by) }{n' f_{XY}(\bx,\by)+n'' f_{X}(\bx)f_{Y}(\by)}\;\rd\bx\rd\by.\end{array}\eeqq
Furthermore, 
\beqq \diy\frac{\bbE\left[\mathfrak{R}'_{n',n''}\right]}{n}+\diy\frac{c_d 2^d}{n}\geq \diy\frac{\bbE\left[\mathfrak{R}'^*_{n',n''}\right]}{n},\eeqq
where $c_d$ is the largest possible degree of any vertex of the MST in $\bbR^d$ as before. Consequently, we have 
\beq\label{eq:final1.ApB} \begin{array}{l} \bigg|\diy\frac{\bbE\left[\mathfrak{R}'_{n',n''}\right]}{n}-\iint \diy\frac{2\alpha\beta f_{XY}(\bx,\by) f_{X}(\bx)f_{Y}(\by)}{\alpha f_{XY}(\bx,\by)+\beta f_{X}(\bx) f_{Y}(\by)}\;\rd\bx\rd\by\bigg|
\leq \mathcal{B}(\alpha,f_{XY},f_{X}f_{Y}),\end{array}\eeq
where $\mathcal{B}$ is defined in (\ref{warpi.func}) and $ \mathcal{A}_{f,n}^{\beta,\alpha}(\bx,\by)$ has been introduced in (\ref{eq:Def:A}). The last line in (\ref{eq:final1.ApB}) is implied from the fact that
\beqq \begin{array}{l}
\diy\sum\limits_{i=1}^M\bbE_{n'_i,n''_i}\Big[(n'_i+n''_i)\; \varsigma_\eta(l,n'_i,n''_i) \Big]
\leq \diy \sum\limits_{i=1}^M O(l)\iint l^{-d/2}\sqrt{b_i}\;\mathcal{A}_{f,n}^{\beta,n'/n}(\bx,\by)\rd\bx \rd\by\\
\\
 \quad+\diy \sum\limits_{i=1}^M O(l) \iint l^d (a_i)^{-1}\diy\frac{2 f_{XY}(\bx,\by)\Big(f_{X}(\bx)f_{Y}(\by)+O(\delta_f\big/(\beta n))\Big)}{n' f_{XY}(\bx,\by)+n'' f_{X}(\bx)f_{Y}(\by)}\;\rd\bx\rd\by.
\end{array}\eeqq
Here $\mathcal{A}_{f,n}^{\beta,n'/n}(\bx,\by)$ is given as (\ref{eq:Def:A}) by substituting $n'/n$ in $\alpha$ such that $\beta=1-\alpha$. Hence, the proof of Theorem \ref{main:Bias} is completed.
\end{proof}
Going back to the proof of (\ref{Optimized:Bias}), without loss of generality assume that $(n)l^{-d}>1$. In the range $d\geq 2$ and $0<\eta\leq 1$. We select $l$ as a function of $n$ and $\beta$ to be the sequence increasing in $n$ which minimizes the maximum of these rates: \beq\label{opt.l} l(n, \beta)=arg\;\min\limits_{l}\max\Big\{l^d(n)^{-\eta/d}, l^{-\eta d}, l^{d}\beta^{-1}n^{-1},c_d 2^d n^{-1} \Big\}. \eeq
 The solution $l=l(n,\beta)$ occurs when $l^d(n)^{-\eta/d}=l^{-\eta d}$, or equivalently $l=\lfloor(n)^{\eta/(d^2(\eta+1))}\rfloor$ and also $l^{d}\beta^{-1}n^{-1}=l^{-\eta d}$ which implies $l=\lfloor(\beta \;n)^{1/(d(1+\eta))}\rfloor$. Substitute this in $l$ in the bound (\ref{bound:Bias}), the RHS expression in (\ref{Optimized:Bias}) for $d\geq 2$ is derived. 
\end{proof}

%----------------------------------------------------------
\subsection{Appendix F}\label{Optimized:alpha}
Our main goal in subsection \ref{subsec:opt.alpha} is to find proportion $\alpha$ such that the parametric MSE rate depending on the joint density $f_{XY}$ and marginal densities $f_X, f_Y$ is minimized. Recalling the explicit bias bound in (\ref{warpi.func}), it can be seen that this function is a complicated function of $f_{XY}$, $f_{X}f_{Y}$ and $\alpha$. By rearranging, we first find an upper bound for $\Delta$ in (\ref{warpi.func}), denoted by $\overline{\Delta}$, as follows:
\beq \label{tighterbound.Delta}\begin{array}{l}\diy  \overline{\Delta}(\alpha,f_{XY},f_{X}f_{Y})
=\diy D(n,l_n,d,\eta)+\widetilde{D}(n,l_n,d)\bbE_{XY}\Big[{G}^{\alpha,\beta}_{f,n}(X,Y)\Big],\end{array}\eeq
where $l_n:=\big\lfloor n^{\eta/(d^2(1+\eta))}\big\rfloor$. From Appendix E we know that optimal $l$ is given in (\ref{opt.l}). 
%minimum  $\big\{l=\lfloor(n)^{\eta/(d^2(\eta+1))}\rfloor, l=\lfloor(\beta \;n)^{1/(d(1+\eta))}\rfloor\big\}$. 
One can check that for $\alpha\leq 1-n^{(\eta/d)-1}$, optimal $l=\big\lfloor n^{\eta/(d^2(1+\eta))}\big\rfloor$ provides a tighter bound. 
In (\ref{tighterbound.Delta}), the constants $D$ and $\overline{D}$ are
\beq \begin{array}{l} D(n,l_n,d,\eta)=\diy c_2l_n^dn^{-1}+c_d2^dn^{-1}+c'l_n^d n^{-\eta/d}+\diy c l_n^d n^{-1/d}+2c_1l_n^{d-1}n^{1/d-1},\end{array}\eeq
\begin{equation}\begin{array}{l} \widetilde{D}(n,l_n,d) 
=\diy 2+n^{-1}2c''\sum\limits_{i=1}^M l_n\; l_n^d a_i^{-1}+n^{-3/2}2 c'_1\sum\limits_{i=1}^M l_n\; l_n^{d/2}\sqrt{b_i}{a_i^2}\\
\qquad+n^{-1}\diy\sum\limits_{i=1}^M 2 n^{-3/2}l_n^{-d/2}\diy\frac{\sqrt{b_i}}{a_i^2}\big(na_il_n^d+n^2a_i^2\big)^{1/2}\big(nb_il_n^d+n^2b_i^2\big)^{1/2}.\end{array}\end{equation}
%Also, introduce $\delta_f$ by 
%\beq\label{def:delta} \delta_f:=\diy\iint\Big|f_{XY}(\bx,\by)-f_{X}(\bx)f_{Y}(\by)\Big|\;\rd\bx \rd\by,\eeq
And the function ${G}^{\alpha,\beta}_{f,n}(\bx,\by)$ is given as the following: 
\begin{equation}\label{eq:C(f,n)} \begin{array}{l} {G}^{\alpha,\beta}_{f,n}(\bx,\by)=
\diy\left(f_{X}(\bx)f_{Y}(\by)+\delta_f\big/(n\beta)\right)\Big(\sqrt{\alpha}f_{XY}(\bx,\by)\\
\\
+\diy\sqrt{\beta}\big(f_{X}(\bx)f_{Y}(\by)+\delta_f\big/(\beta n)\Big)
\Big/\left(\alpha f_{XY}(\bx,\by)+\beta f_{X}(\bx)f_{Y}(\by)\right)^2,\end{array}\end{equation}
where $\delta_f$ is given in (\ref{delta.def}). After all still the expression (\ref{eq:C(f,n)}) is complicated to optimize therefore we use the fact that $0\leq \alpha,\beta\leq 1$ and bound the function $\mathbf{G}^{\alpha,\beta}_{f,n}(\bx,\by)$. Define set $\Gamma$ as 
\beqq \Gamma:=\Big\{\epsilon_{XY}:\big|\epsilon_{XY}(\bt)- \epsilon_{XY}(\bt')\big|\leq \overline{K}\|\bt-\bt'\|_d^\eta\Big\},\eeqq
where  
$$\overline{K}= C^U_\epsilon K\; \left\{C^L_{XY}+C^L_{X}+C^L_YC^L_XC^U_{X}\right\}.$$
Here $K$ is the smoothness constant. Notice the set $\Gamma$ is a convex set. We bound $\overline{\Delta}$ by

\begin{equation}\label{def:varpi} 
 \diy\widetilde{\Delta}(\alpha,\epsilon_{XY})=\diy D(n,l_n,d,\eta)+\widetilde{D}(n,l_n,d)\; C^U_{XY}\diy\iint_{\bbS_{XY}} \widetilde{{G}}^{\alpha,\beta}_{\epsilon_{XY},n}(\bx,\by)\;\rd\bx\rd\by.
\end{equation}
 Set $C_n=C^L_{XY}\; n/2$,
\begin{equation}\label{widetildeG}
\widetilde{{G}}^{\alpha,\beta}_{n}(\epsilon_{XY})=\diy\frac{\left(\epsilon_{XY}^{-1}(\bx,\by)+(\beta C_n)^{-1})(1+\epsilon_{XY}^{-1}(\bx\by)+(\beta C_n)^{-1}\right)}{\left(\alpha+\beta \epsilon_{XY}^{-1}(\bx,\by)\right)^2}.
\end{equation}
We simplify (\ref{widetildeG}) by
\begin{equation}\label{widetildeG}
\widetilde{{G}}^{\alpha,\beta}_{n}(\epsilon_{XY})=\diy\frac{\Big(1+(\beta C_n)^{-1}\epsilon_{XY}\Big)\Big(1+\epsilon_{XY}+(\beta C_n)^{-1}\epsilon_{XY}\Big)}{\Big(\alpha \epsilon_{XY}+\beta \Big)^2}.
\end{equation}
Under condition 
\begin{equation}\label{con1}\diy\frac{2}{C_n}\leq \alpha \leq \min\left\{\diy\frac{1}{2}+\frac{1}{2C_n},\frac{1}{3}+\frac{2}{3C_n}\right\},\end{equation} 
$\widetilde{{G}}^{\alpha,\beta}_{n}(\epsilon_{XY})$ is an increasing function in $\epsilon$. Furthermore, for $\alpha\leq \diy\frac{1}{4}$ and 
\beq\label{con2} C^L_\epsilon\leq \epsilon_{XY}\leq\min\big\{C_\epsilon^U,\theta^U(\alpha)\big\},\;\;\; \hbox{where}\;\;\theta^U(\alpha)=\frac{1-4\alpha+1/C_n}{2\alpha} ,\eeq
the function $\widetilde{{G}}^{\alpha,\beta}_{n}(\epsilon_{XY})$ is strictly concave. 
Next, to find optimal $\alpha$ we consider the following optimization problem:
\def\bt{\mathbf{t}}
\begin{equation}\begin{aligned}
& \underset{\alpha}{\text{min}}\;\; \underset{\epsilon_{XY}\in \Gamma}{\text{max}}
& & \widetilde{\Delta}(\alpha,\epsilon_{XY})+c_d(1-\alpha)n^{-1}\\
& \text{subject to}
& & C^L_{\epsilon}\leq  \epsilon_{XY}\leq C^U_{\epsilon},
\end{aligned}\end{equation}
here $\epsilon_{XY}=f_{XY}\big/f_{X}f_{Y}$, $C^U_{\epsilon}=C^U_{XY}\big/C^L_{X}C^L_{Y}$ and $C^L_{\epsilon}=C^L_{XY}\big/C^U_{X}C^U_{Y}$, such that $C^L_\epsilon\leq 1$. We know that under conditions (\ref{con1}) and (\ref{con2}),
%$$\diy\frac{2}{C_n}\leq \alpha\leq \diy\frac{1}{4}+\diy\frac{1}{4C_n},$$
%and 
%\beqq \epsilon_{XY}\leq\frac{1-4\alpha+1/C_n}{2\alpha} :=\theta^L(\alpha),\eeqq
the function $\widetilde{{G}}^{\alpha,\beta}_{n}$ is strictly concave and increasing in $\epsilon_{XY}$. 
%Note that when sample size $n$ is large enough then $\alpha\in(0,\diy\frac{1}{4}]$ and $\epsilon_{XY}\leq \diy\frac{1-4\alpha}{2\alpha}$. 
%Note that for $\alpha\geq \diy\frac{1}{4}+\frac{1}{4C_n}$, we let $\theta^U(\alpha)=0$.
\def\ep{\epsilon}
We first solve the optimization problem:
\begin{equation}\begin{aligned}
&\underset{\epsilon_{XY}\in \Gamma}{\text{max}}
& & \diy\iint_{\bbS_{XY}} \widetilde{{G}}^{\alpha,\beta}_{n}\big(\epsilon_{XY}(\bx,\by)\big)\;\rd\bx\rd\by\\
& \text{subject to}
& & \theta^L_{\ep}(\alpha)\bbV(\bbS_{XY})\leq \iint_{\bbS_{XY}} \epsilon_{XY}(\bx,\by)\rd\bx\rd\by\\
&
& & \qquad \leq \theta^U_{\ep}(\alpha) \bbV(\bbS_{XY}),
\end{aligned}\end{equation}
where
\beq\label{eps:thetaLU} \theta^L_{\ep}(\alpha):= C^L_\epsilon,\;\;\;\;
\theta^U_{\ep}(\alpha):=\min\{C^U_\epsilon,\theta^U(\alpha)\}.
\eeq
The Lagrangian for this problem is 
\beqq \begin{array}{l}L(\epsilon_{XY},\lambda_1,\lambda_2)=\diy\iint_{\bbS_{XY}} \widetilde{{G}}^{\alpha,\beta}_{n}\big(\epsilon_{XY}(\bx,\by)\big)\;\rd\bx\rd\by
-\lambda_1\left(\diy\iint_{\bbS_{XY}}\epsilon_{XY}(\bx,\by) \;\rd\bx\rd\by- \theta^U_{\epsilon}(\alpha)\bbV(\bbS_{XY})\right)\\
\\
\qquad\qquad\qquad\qquad-\lambda_2\left( \theta^L_{\epsilon}(\alpha)\bbV(\bbS_{XY})-\diy\iint_{\bbS_{XY}}\epsilon_{XY}(\bx,\by)\;\rd\bx\rd\by\right).\end{array}\eeqq
In this case the optimum $\epsilon^*_{XY}$ is bounded, $\theta^L_\ep(\alpha)\leq \epsilon^*_{XY}\leq \theta^U_\ep(\alpha),$
and Lagrangian multiplier $\lambda^*_1,\;\lambda^*_2\geq 0$ such that
\beqq \underset{\lambda_1,\lambda_2\geq 0}{\text{min}}\;\; \underset{\epsilon_{XY}\in \Gamma}{\text{max}}\; L(\epsilon_{XY},\lambda_1,\lambda_2)=L(\epsilon^*_{XY},\lambda^*_1,\lambda^*_2).\eeqq
Set
$G'_{n}(\epsilon_{XY})=\diy\frac{\rd}{\rd \epsilon_{XY}} \widetilde{{G}}^{\alpha,\beta}_{n}(\epsilon_{XY}).$ In view of the concavity of $\widetilde{{G}}^{\alpha,\beta}_{\epsilon_{XY},n}$ and Lemma 1, page 227 in \cite{Lu}, maximizing $L(\epsilon_{XY},\lambda^*_1,\lambda^*_2)$ over $\epsilon_{XY}$ is equivalent to 
 \beq\label{eq:opt1.1} \diy \iint_{\bbS_{XY}}\Big\{G'_{n}\big(\epsilon^*_{XY}(\bx,\by)\big)-(\lambda^*_1-\lambda^*_2)\Big\}\epsilon_{XY}(\bx,\by) \;\rd\bx\rd\by\leq 0,\eeq
for all $\theta_\ep^L(\alpha)\leq \epsilon^*_{XY}\leq \theta_\ep^U(\alpha)$, and 
\beq \label{eq:opt1.2}  \diy \iint_{\bbS_{XY}}\Big\{G'_{n}\big(\epsilon^*_{XY}(\bx,\by)\big)-(\lambda^*_1-\lambda^*_2)\Big\}\epsilon^*_{XY}(\bx,\by) \;\rd\bx\rd\by = 0.\eeq
Denote  ${G'}^{-1}_{n}$ the inverse function of $G'_{n}$. Since $G'_{n}$ is strictly decreasing in $\epsilon^*_{XY}$ (this is because $\tilde{G}^{\alpha,\beta}_n(\epsilon_{XY})$ is strictly concave, then ${G'}^{-1}_{n}$ is continues and strictly decreasing in $\epsilon^*_{XY}$). From (\ref{eq:opt1.1}) and (\ref{eq:opt1.2}), we see immediately that on any interval $\theta^L_{\epsilon}(\alpha)\leq \epsilon^*_{XY}\leq \theta^U_{\epsilon}(\alpha)$, we have $\epsilon^*_{XY}={G'}^{-1}_{n}(\lambda^*_1-\lambda^*_2)$. We can write then 
$$G'_{n}\big(\theta^U_{\epsilon}(\alpha)\big)\leq \lambda^*_1-\lambda^*_2\leq G'_{n}\big(\theta^L_{\epsilon}(\alpha)\big),$$
and $\lambda^*_1,\lambda^*_2\geq 0$. Next, we intend to find the solution of 
\beqq \underset{\lambda_1,\lambda_2\geq 0}{\text{min}}\;\;\overline{{G}}^{\alpha,\beta}_{n}(\lam_1,\lam_2),\;\;\; \hbox{where}\eeqq
\beqq \begin{array}{l}\overline{{G}}^{\alpha,\beta}_{n}(\lam_1,\lam_2)=\bbV(\bbS_{XY})\bigg\{ \widetilde{{G}}^{\alpha,\beta}_{n}\big({G'}^{-1}_{n}(\lambda_1-\lambda_2)\big)-(\lam_1-\lam_2){G'}^{-1}_{n}(\lambda_1-\lambda_2)+\lam_1 \theta^U_\epsilon(\alpha)-\lam_2 \theta^L_\epsilon(\alpha)\bigg\}.
\end{array}\eeqq
The function $\overline{{G}}^{\alpha,\beta}_{n}(\lam_1,\lam_2)$ is increasing in $\lambda_1$ and $\lambda_2$, therefore it takes its minimum  at $(\lam^*_1,\lam^*_2)=(G'_n \big({\theta^U_\epsilon}(\alpha)\big),0)$. This implies that $\epsilon^*_{XY}=\theta^U_\epsilon(\alpha)$. We continue by going back to our primary minimization over $\alpha$:
\begin{equation}\label{const0:2}\begin{aligned}
& \underset{\alpha}{\text{min\;\;}}
& & \widetilde{\Delta}(\alpha,\epsilon^*_{XY})+c_d(1-\alpha)n^{-1}\\
& \text{subject to}
& & \alpha^L_0\leq  \alpha \leq \alpha^U_0,
\end{aligned}\end{equation}
where $\alpha_0^L=\diy\frac{2}{C_n}$ and $\alpha_0^U=\min\Big\{\diy\frac{1}{4}, 1-n^{\eta/d-1}\Big\}$.
 We know that  $\diy\frac{1}{4}\leq \frac{1}{3}+\diy\frac{2}{3C_n}$ and $\diy\frac{1}{4}\leq \frac{1}{2}+\frac{1}{2C_n}$, therefore the condition below
$$\diy\frac{2}{C_n}\leq \alpha\leq \min\Big\{\diy\frac{1}{4}, 1-n^{\eta/d-1}\Big\},$$
implies previous conditions on $\alpha$. Since the objective function (\ref{const0:2}) is a complicated function in $\alpha$, it is not feasible to determine whether it is a convex function in $\alpha$. For this reason let us solve the optimization problem in (\ref{const0:2}) in a special case when $C^U_\ep\leq \theta^U(\alpha)$. This implies $\ep^*_{XY}=C^U_\ep$. Under assumption $C^U_\ep$ the objective function in (\ref{const0:2}) is convex in $\alpha$. Also the case $C^U_\ep\leq \theta^U(\alpha)$ is equivalent to $\alpha\leq \diy\frac{1+1/C_n}{4+2C^U_\ep}$. Therefore in the optimization problem we have constraint 
\begin{equation*}\diy\frac{2}{C_n}\leq \alpha\leq \min\Big\{\diy\frac{1}{4}, \diy\frac{1+1/C_n}{4+2C^U_\ep}, 1-n^{\eta/d-1}\Big\}.\end{equation*}
We know that $\widetilde{\Delta}(\alpha,\epsilon^*_{XY})+c_d(1-\alpha)n^{-1}$ is convex over $\alpha\in[\alpha^L_0,\alpha^U_0]$. So, the problem becomes ordinary convex optimization problem. Let $\tilde{\alpha}$, $\tilde{\lambda}_1$ and $\tilde{\lambda}_2$ be any points that satisfy the KKT conditions for this problem:
\beq\label{KKT} \begin{array}{l}\alpha_0^L-\tilde{\alpha} \leq 0,\;\;\; \tilde{\alpha}-\alpha_0^U\leq 0,\;\;\; \tilde{\lambda}_1,\tilde{\lambda}_2\geq 0,\\
\\
\tilde{\lambda}_1(\alpha_0^L-\tilde{\alpha})=0,\;\;\; \tilde{\lambda}_2(\tilde{\alpha}-\alpha_0^U)=0,\\
\\
\diy\frac{\rd}{\rd \alpha} \Big(\widetilde{\Delta}(\tilde{\alpha},\epsilon^*_{XY})+c_d(1-\tilde{\alpha})\;n^{-1}\Big)-\tilde{\lambda}_1+\tilde{\lambda}_2=0.
\end{array}\eeq
Recall $\Xi(\alpha)$ from (\ref{def.Xi}):
\beqq\begin{array}{l} \Xi(\alpha)=\diy\frac{\rd}{\rd \alpha} \Big(\widetilde{\Delta}({\alpha},\epsilon^*_{XY})+c_d(1-\alpha)n^{-1}\Big),
\end{array}\eeqq
where $\widetilde{\Delta}$ is given in (\ref{def:varpi}). 
So the last condition in (\ref{KKT}) becomes $\Xi(\tilde{\alpha})=\tilde{\lambda}_1-\tilde{\lambda}_2$. We then have
\beqq \alpha_0^L\leq \Xi^{-1}(\tilde{\lambda}_1-\tilde{\lambda}_2)\leq \alpha_0^U,
\eeqq
where $\Xi^{-1}$ is inverse function of $\Xi$. Since $\alpha_0^L\neq\alpha_0^U$, at least one of $\tilde{\lambda}_1$ or $\tilde{\lambda}_2$ should be zero:
\begin{itemize}
\item $\tilde{\lambda}_1=0$, $\tilde{\lambda}_2\neq 0$. Then $\tilde{\alpha}=\alpha_0^U$ and implies $\tilde{\lambda}_2=-\Xi(\alpha_0^U)$. Since $\tilde{\lambda}_2>0$, so this leads to $\Xi(\alpha_0^U)<0$. 
\item $\tilde{\lambda}_2=0$, $\tilde{\lambda}_1\neq 0$. Then $\tilde{\alpha}=\alpha_0^L$ and implies $\tilde{\lambda}_1=\Xi(\alpha_0^L)$. We know that $\tilde{\lambda}_1>0$, hence $\Xi(\alpha_0^L)>0$. 
\item  $\tilde{\lambda}_1=0$, $\tilde{\lambda}_2=0$. Then $\tilde{\alpha}=\Xi^{-1}(0)$ and so $\alpha_0^L\leq \Xi^{-1}(0)\leq \alpha_0^U$.
\end{itemize}
Consequently, by following the behavior of $\Xi(\alpha)$ with respect to $\alpha^L_0$ and $\alpha^U_0$, we are able to detect optimal $\tilde{\alpha}$, $\tilde{\lambda}_1$ and $\tilde{\lambda}_2$. For instance, if $\Xi(\alpha)$ is positive for all $\alpha\in[\alpha^L_0, \alpha^U_0]$ then we conclude that $\tilde{\alpha}=\alpha^L_0$.

%%%%%%%%%%%%%%%%%%%%%%%%%%%%%%%%%%%%%%%%%%

\reftitle{References}
%\label{sec:refs}
\bibliographystyle{IEEEbib}
\bibliography{refs}
%\begin{thebibliography}{999}
%\input{Refrence.tex}
%\end{thebibliography}
\end{document}